\documentclass{article}

\usepackage[preprint]{neurips_2026}

\usepackage[utf8]{inputenc} 
\usepackage[T1]{fontenc}    
\usepackage{hyperref}       
\usepackage{url}            
\usepackage{booktabs}       
\usepackage{amsfonts}       
\usepackage{nicefrac}       
\usepackage{microtype}      
\usepackage{xcolor}         
\usepackage{multirow}    
\usepackage{array}       
\usepackage{caption}     
\usepackage{tabularx}   
\usepackage{graphicx}
\usepackage{xspace}
\usepackage{color}
\usepackage{tabularray}
\UseTblrLibrary{booktabs}
\usepackage{enumitem}
\usepackage{rotating}
\usepackage{longtable}
\usepackage{etoc}

\emergencystretch=\maxdimen
\newcommand*{\benchmarkname}{SpeechDx\xspace}

\newcolumntype{Y}[1]{>{\hsize=#1\hsize\raggedright\arraybackslash}X}

\newenvironment{s_itemize}{
\begin{itemize}[nosep,leftmargin=20pt]
}{\end{itemize}}

\title{\benchmarkname: A Multi-Task Benchmark for Clinical Speech AI}

\author{%
  Sejal Bhalla, Larry Kieu, Aina Merchant, Eyal de Lara, Alex Mariakakis \\[0.5em]
  University of Toronto, Canada \\
  \texttt{sejal@cs.toronto.edu}
}
\begin{document}

\maketitle

\begin{abstract}
  Speech offers a uniquely informative window into health by simultaneously engaging neurological, motor, respiratory, and vocal systems.
Current clinical speech AI methods have largely progressed through isolated condition-specific studies, making results difficult to compare and generalization difficult to assess.
We introduce \benchmarkname, a large-scale benchmark for clinical speech AI spanning 12 datasets and 27 tasks across diverse health conditions. 
To enable evaluation across shared clinical mechanisms, \benchmarkname structures tasks by the stage of speech production they disrupt: conceptualization, formulation, and articulation. 
The benchmark tests generalization by including tasks with limited labeled data and evaluating the same health condition across multiple datasets, distinguishing clinically meaningful patterns from dataset artefacts.
We systematically evaluate 12 state-of-the-art audio encoders across all tasks and under zero-shot cross-condition transfer.
Results show that large-scale speech models represent the strongest overall baselines, domain-specific models improve performance only on closely matched tasks, and no current representation generalizes reliably across the clinical speech landscape.
\benchmarkname establishes a shared evaluation framework for tracking progress toward general-purpose clinical speech representations.

\end{abstract}

\section{Introduction}
Speech has been widely studied as a digital biomarker of human health. Its production requires the coordinated action of the respiratory, vocal, neurological, and cognitive systems, so disruptions to any of these systems due to disease or dysfunction leave measurable traces in the voice. Unlike many clinical assessment tools, speech can be captured non-invasively, remotely, and at negligible cost, making it particularly well-suited for continuous monitoring, population-level screening, and longitudinal disease tracking. Over the past few decades, this has motivated a substantial body of research exploring the automation of speech analysis for a wide range of conditions such as COVID-19 \cite{despotovic2021covid, usman2022speechcovid}, dysarthria~\cite{shih2022dysarthria, riosurrego2023dysarthria, JAVANMARDIdysarthria}, Parkinson's disease~\cite{reddy2023parkinson, khaskhoussy2023parkinson, morovelazquez2021parkinson}, Alzheimer's disease~\cite{zolnoori2023alzheimers, martineznicolas2021alzheimer, braun2023dementia}, depression~\cite{wu2023depression, koops2023depression, cummins2015depression}, and vocal pathologies~\cite{liu2024voicepathology, koudounas25b_voicepathology}. Dedicated challenges further demonstrate the potential of deep learning for health assessment from speech~\cite{ksofc_challenge, gale2022psst_challenge, luz20_address_challenge, speechwellnesschallenge, muguli2021dicovachallenge, luz2023multilingualalzheimerschallenge}.

Despite this potential, very few systems have achieved real-world clinical deployment. A recurring obstacle is the fragmented nature of the field. Research has progressed in silos, with models trained and evaluated on individual datasets under inconsistent protocols, limiting generalizability across conditions. Even within a single condition, models trained on small controlled corpora consistently fail to generalize to unseen data. This failure is largely attributed to models learning spurious correlations from confounding factors in the data — differences in recording conditions, demographic composition, and acquisition hardware — rather than the underlying clinical signal~\cite{speech_distribution_shift, torgo_artefacts, covid_biases}. The field lacks a standardized evaluation framework for quantifying progress, assessing generalization across datasets and conditions, and identifying robust modeling approaches.

We address this gap by introducing \benchmarkname, a clinical speech AI benchmark comprising \textbf{12 publicly available} speech datasets with \textbf{27 tasks} spanning \textbf{nine health and affective conditions}. One of \benchmarkname's distinguishing features is its organization around the speech production process. Following the framework proposed by Berisha and Liss~\cite{clinical_taxonomy}, we classify conditions and tasks according to the stage of speech production at which the condition exerts its primary effect: (1) conceptualization, when a communicative intention is formed; (2) formulation, when that intention is encoded into linguistic and phonological structure; and (3) articulation, where respiratory and motor systems execute the spoken output. Diverse health conditions disrupt different stages of this cascade, producing distinct types of acoustic irregularity.
 
We use \benchmarkname to evaluate 12 state-of-the-art audio and speech encoders representing different representation learning paradigms. We first benchmark the models across all tasks, establishing how well the pretrained representations encode clinically relevant information. Second, we conduct a zero-shot cross-condition analysis, revealing shared acoustic structure across conditions and identifying where generalization breaks down. To summarize, our contributions are as follows: 

\begin{s_itemize}
    \item We introduce \benchmarkname, the first large-scale benchmark specifically designed to advance clinical speech AI. This benchmark comprises 12 datasets and 27 tasks across nine health conditions. The codebase is available at \url{https://anonymous.4open.science/r/SpeechDx-F584}.
    \item We provide a systematic evaluation of 12 state-of-the-art audio encoders, establishing a standardized and reproducible baseline for performance comparison across the clinical domains.
    \item We conduct a zero-shot cross-condition transfer analysis to reveal which conditions share learnable acoustic structure and where transfer degrades.
\end{s_itemize}
\section{Background and Related Work}
\subsection{Speech as a Biomarker}
Deviations in vocal attributes such as pitch, intensity, resonance, and temporal structure constitute measurable markers of disordered voice production. Such deviations arise from perturbations in the biological systems underlying speech generation~\cite{verdolini2006voice, fagherazzi2021voice}. This sensitivity is fundamental to the utility of speech as a biomarker \cite{robin_evaluation} and contributes to its recognition as a vital sign \cite{speech_vital_sign}. 
Its non-invasive and low-cost acquisition further enables repeated and longitudinal measurements that support applications in screening, monitoring, and prognosis.

Berisha and Liss~\cite{clinical_taxonomy} propose a framework that organizes health conditions by how directly they disrupt the speech production mechanism. At one end of the spectrum, conditions directly impact the mechanics of acoustic speech production. On the other end, conditions disrupt cognitive-affective processes or linguistic planning, resulting in changes to speech content and formulation. \benchmarkname adopts this framework as the basis for task selection and organization.

\subsection{Clinical Speech AI}
The field of clinical speech AI broadly studies computational methods that extract acoustic, prosodic, and linguistic information from speech to infer clinically relevant outcomes \cite{clinical_speech_AI}. Early work in the field relied on hand-engineered features such as MFCCs, jitter, shimmer, and prosodic descriptors~\cite{opensmile}. These representations have proven to be useful for tasks spanning COVID-19 detection \cite{covid_features}, Parkinson's disease classification \cite{parkinsons_features}, and depression screening \cite{depression_features}, among others~\cite{copd_features, parkinsons_dysarthria_features, cancer_features}. More recent work has shifted toward deep neural networks trained directly on audio waveforms or spectrograms~\cite{speech_dl, ad_dl, vp_dl, covid_dl, aphasia_dl, depression_dl}. However, both conventional and deep approaches are typically optimized for individual conditions. Although these models often achieve strong within-dataset performance, they frequently fail under distribution shift \cite{speech_distribution_shift, clinical_speech_AI, covid_distribution_shift}.

Innovations in self-supervised learning across vision~\cite{mae, dinov2, jepa_vision} and language~\cite{bert, openai, llama} have shown that it is possible to learn universal representations that transfer across several downstream tasks with minimal adaptation. An analogous capability is desirable in clinical speech AI since labeled datasets are small and costly to collect. A disease-agnostic representation could provide a shared basis for learning across multiple clinical tasks, reducing dependence on large labeled datasets for each disease. Initial studies using representations from models such as wav2vec~2.0~\cite{wav2vec2}, HuBERT~\cite{hubert}, and WavLM~\cite{wavlm} have shown promising results on clinical speech tasks~\cite{w2v2_dysarthria, whisper_aphasia, ssl_depression, w2v2_emotion, whisper_dysarthria, pre-trained_covid}, suggesting that pretrained speech encoders capture information relevant to health assessment.

Recent work has therefore shifted toward health-oriented audio representations. WavRx~\cite{wavrx} extended WavLM with a modulation dynamics module to capture respiration and articulation abnormalities, while HeAR~\cite{hear} included a masked autoencoder trained on 313 million health acoustic clips. Both represent important steps towards general audio representations for health. However, their evaluation scope remains limited: WavRx covers six datasets and four pathologies, while HeAR was tested on tasks spanning respiratory sounds like coughing and breathing. The broader scope of speech-affecting diseases spanning motor, cognitive-linguistic, and affective conditions remains largely unaddressed. 

Standardized benchmarks have driven systematic progress in adjacent domains; SUPERB~\cite{superb} established a shared multi-task evaluation protocol that accelerated semantic speech research, while HEAR~\cite{hear_acoustic} provided a framework that advanced general audio representation learning. \benchmarkname serves this purpose for clinical speech AI, providing a catalyst for the development of general-purpose health audio representations that transfer across clinical tasks and populations.

\section{Datasets and Tasks}
\label{sec:tasks-and-datasets}

Subject to clinical relevance and data access constraints, the dataset curation process yielded 12 datasets with 27 downstream tasks.
\autoref{tab:benchmark_tasks} summarizes the key characteristics of these datasets and tasks, but more detailed descriptions can be found in Appendices~\ref{app:datasets} and \ref{app:tasks} respectively.
Below, we describe how the datasets are positioned under Berisha and Liss' framework for speech production~\cite{clinical_taxonomy}.

\begin{table*}[!ht]
\centering
\caption{The datasets and tasks comprising \benchmarkname. Task types are either classification~(C), multi-label classification~(M), or regression~(R). Dataset splits either entail a single train-validation-test split (TVT) or 5-fold subject-wise cross-validation (5-fold). Sample sizes denote the number of recordings, with the number of unique subjects shown in parentheses.}
\label{tab:benchmark_tasks}
\scriptsize
\setlength{\tabcolsep}{2pt}
\resizebox{\textwidth}{!}{
\begin{tabular}{
>{\raggedright\arraybackslash}p{2cm}
>{\raggedright\arraybackslash}p{1.5cm}
>{\centering\arraybackslash}p{0.45cm}
>{\raggedright\arraybackslash}p{2.7cm}
>{\centering\arraybackslash}p{0.5cm}
>{\raggedright\arraybackslash}p{0.7cm}
>{\raggedright\arraybackslash}p{3.8cm}
}

\toprule
\textbf{Category} & \textbf{Dataset} & \textbf{ID} & \textbf{Task} & \textbf{Type} & \textbf{Split} & \textbf{Samples (Subjects)} \\
\midrule

\multirow{6}{=}{\textbf{Conceptualization}}
& \multirow{2}{=}{EDAIC-WOZ \cite{edaic}}
& T1 & Depression / healthy & C & TVT & 163 (163) / 56 (56) / 56 (56) \\
& & T2 & PHQ-8 score & R & TVT & 163~(163) / 56~(56) / 56~(56) \\

\cmidrule{2-7}
& \multirow{2}{=}{RAVDESS \cite{ravdess}}
& T3 & Emotion classification & C & 5-fold & 1,140~(19) / 300~(5) \\
& & T4 & Negative / non-negative emotion & C & 5-fold & 1,140~(19) / 300~(5) \\

\cmidrule{2-7}
& \multirow{2}{=}{IEMOCAP \cite{iemocap}}
& T5 & Emotion classification & C & 5-fold & 5,843~(8) / 1,537~(2) \\
& & T6 & Negative / non-negative emotion & C & 5-fold & 5,843~(8) / 1,537~(2) \\

\midrule

\multirow{3}{=}{\textbf{Formulation}}
& \multirow{2}{=}{DementiaBank \cite{address-M}}
& T7 & Dementia / healthy & C & TVT & 237 (237) / 8 (8) / 46 (46) \\
& & T8 & MMSE score & R & TVT & 236~(236) / 8~(8) / 46~(46) \\

\cmidrule{2-7}
& AphasiaBank \cite{aphasiabank}
& T9 & Aphasia / healthy & C & TVT & 740~(143) / 160~(21) / 237~(42) \\

\midrule

\multirow{9}{=}{\textbf{Articulation (Neuromuscular)}}
& \multirow{2}{=}{TORGO \cite{torgo}}
& T10 & Dysarthria / healthy & C & 5-fold & 7,605 (12) / 1,811 (3) \\
& & T11 & Dysarthria severity & R & 5-fold & 2,379 (6) / 802 (2) \\

\cmidrule{2-7}
& UASpeech \cite{uaspeech}
& T12 & Dysarthria / healthy & C & 5-fold & 16,830 (22) / 4,590 (6) \\

\cmidrule{2-7}
& \multirow{4}{=}{MDVR-KCL \cite{mdvrkcl}}
& T13 & Parkinson's / healthy & C & 5-fold & 59 (30) / 14 (7) \\
& & T14 & UPDRS-5 score & R & 5-fold & 59 (30) / 14 (7) \\
& & T15 & UPDRS-18 score & R & 5-fold & 59 (30) / 14 (7) \\
& & T16 & H\&Y score & R & 5-fold & 59 (30) / 14 (7) \\

\cmidrule{2-7}
& \multirow{2}{=}{KSoF-C \cite{ksof}}
& T17 & Disfluency / healthy & C & 5-fold & 4,307 (31) / 1,036 (6) \\
& & T18 & Disfluency classification & M & 5-fold & 4,394 (30) / 1,203 (7) \\

\midrule

\multirow{9}{=}{\textbf{Articulation (Phonatory / Respiratory)}}
& \multirow{5}{=}{COVID-19 Sounds \cite{covid19sounds}}
& T19 & Symptomatic / healthy (official~subset) & C & TVT & 6,565~(4,635) / 943~(663) / 1,948~(1,325) \\
& & T20 & Symptomatic / healthy & C & TVT & 37,353~(25,252) / 5,271~(3,608) / 10,761~(7,216) \\
& & T21 & COVID-19 / non-COVID-19 (official~subset) & C & TVT & 1,017~(700) / 141~(100) / 324~(200) \\
& & T22 & COVID-19 / non-COVID-19 & C & TVT & 5,606~(3,711) / 804~(531) / 1,609~(1,061) \\
& & T23 & Symptom classification & M & TVT & 19,541~(15,459) / 2,914~(2,209) / 5,591~(4,418) \\

\cmidrule{2-7}
& \multirow{3}{=}{Coswara \cite{coswara}}
& T24 & Symptomatic / healthy & C & TVT & 3,786~(1,896) / 542~(271) / 1,083~(542) \\
& & T25 & COVID-19 / non-COVID-19 & C & TVT & 2,915~(1,459) / 418~(209) / 834~(418) \\
& & T26 & Symptom classification & M & TVT & 1,459~(731) / 210~(105) / 417~(209) \\

\cmidrule{2-7}
& AVFAD \cite{avfad}
& T27 & Vocal pathology / healthy & C & TVT & 3,968~(496) / 568~(71) / 1,135~(142) \\

\bottomrule
\end{tabular}
}
\vspace{-3em}
\end{table*}

\subsection{Conceptualization Disorders}
Conceptualization is the first stage of speech production, where the speaker forms a communicative intention and decides what to express. This stage is driven by cognitive and affective processes --- how the speaker feels, what they want to communicate, and how much they elaborate. Disruptions here alter cognitive drivers that result in subtle acoustic changes such as reduced speaking rate, flattened pitch variation, truncated responses, and altered patterns of emphasis and timing. 

Depression and emotional dysregulation are among the most common conditions affecting this stage. Depression reduces psychomotor drive and cognitive engagement, while emotional state more broadly shapes the prosodic and temporal properties of spoken output. Both are clinically significant conditions for which scalable, low-cost assessment tools are actively needed. We use EDAIC-WOZ~\cite{edaic} to probe the presence and severity of depression, and RAVDESS~\cite{ravdess} and IEMOCAP~\cite{iemocap} to probe multi-class emotion recognition in acted and naturalistic speech, respectively.
 
\subsection{Formulation Disorders}
The formulation stage involves encoding communicative intent into linguistic structures through word selection, syntactical rules, and phonetic sequencing. Disrupted formulation often results in speech that is acoustically fluent yet contains unusual sentence structures or phonemic substitutions. 

Language disorders following neurological damage and neurodegenerative disease are the most common disruptors of this stage. Aphasia, arising from stroke or focal brain injury, directly impairs lexical retrieval and syntactic encoding. On the other hand, Alzheimer's disease progressively disrupts semantic memory and word access, with downstream effects on the coherence of spoken output. These conditions represent two of the most clinically prevalent acquired communication disorders in adults, and detecting them from speech alone is highly valuable for tracking disease progression. We use standardized discourse recordings from AphasiaBank~\cite{aphasiabank} to detect the presence of aphasia and DementiaBank~\cite{address-M} to detect the presence and severity of Alzheimer's disease.
 

\subsection{Articulation Disorders}
Articulation is the stage at which planned speech is executed through coordinated movement of the respiratory, laryngeal, and articulatory systems. This execution can be broadly decomposed into two subsystems: (1) the neuromuscular articulatory subsystem, which modulates airflow into intelligible speech, and (2) the phonatory and respiratory subsystem, which generates airflow in the first place. 

\textbf{Neuromuscular Disorders.} These disorders reduce the precision, speed, and coordination of the vocal articulators. This reduction results in speech that is imprecise and less intelligible. We represent this class of disorders with two examples: (1) dysarthria, a group of motor speech disorders characterized by weakness or incoordination of the speech musculature; and (2) disfluency, which disrupts the timing and sequencing of speech production. 
TORGO~\cite{torgo} and UASpeech~\cite{uaspeech} support dysarthria detection and severity estimation among speakers with diverse etiology. MDVR-KCL~\cite{mdvrkcl} targets hypokinetic dysarthria as observed in Parkinson's disease, which is characterized by reduced vocal loudness, monotone pitch, and imprecise articulation. Finally, KSoF-C~\cite{ksof} provides fluency-labeled recordings to enable disfluency classification and stuttering detection.
 
\textbf{Phonatory and Respiratory Disorders.} The respiratory and phonatory subsystem is responsible for the airflow and vocal fold vibration that underlie all voiced sounds. Conditions affecting this subsystem disrupt speech at its source, producing what could theoretically be the most directly measurable acoustic deviations. We consider both disease detection and symptom characterization in this category of disorders. The latter is particularly relevant for respiratory conditions that are accompanied by cough, cold, and breathing difficulty, each of which exhibits distinct acoustic patterns during speech.
We use COVID-19 Sounds~\cite{covid19sounds} and Coswara~\cite{coswara} for respiratory symptom characterization and COVID-19 detection, and AVFAD~\cite{avfad} for vocal pathology detection.
\section{Methods}
\label{sec:methods}

\subsection{Audio Preprocessing}
All audio is resampled to 16\,kHz, mono-channeled, and normalized prior to modeling. To accommodate variable-length recordings, inputs shorter than a model's minimum accepted length are zero-padded, and those exceeding the maximum accepted length are divided into non-overlapping chunks whose embeddings are mean-pooled to produce a single representation. 

\subsection{Models}
We evaluate 12 state-of-the-art audio and speech models that have been widely used for spoken language, paralinguistic, and health-related tasks. These models are selected to cover a range of training data sources (general speech, general audio, and domain-specific audio), supervision paradigms, and learning objectives, enabling comparisons across varying levels of data scale and domain alignment. Model details are provided in \autoref{tab:model_training_data}, with expanded descriptions in Appendix \ref{app:models}.

\textbf{Speech Models.} Following prior work demonstrating the use of self-supervised speech representations in clinical and paralinguistic applications~\cite{w2v2_dysarthria, w2v2_emotion, pre-trained_covid, pre-trained_pathological, pre-trained_voice_disorder, ssl_depression}, we evaluate wav2vec~2.0~\cite{wav2vec2}, HuBERT~\cite{hubert}, WavLM~\cite{wavlm}, and MMS~\cite{mms}. These models utilize large-scale self-supervised speech pretraining followed by supervised fine-tuning for automatic speech recognition, but they differ in their pretraining objectives and data. wav2vec~2.0 uses contrastive learning over quantized representations, HuBERT uses masked prediction with offline cluster targets, WavLM adds a denoising objective, and MMS extends this framework to over 1,400 languages. We additionally include the Qwen3-TTS-Tokenizer~\cite{Qwen3}, a neural audio codec trained on multilingual speech corpora with a self-supervised reconstruction objective; and Whisper~\cite{whisper}, a fully supervised model trained on weakly labeled speech via sequence-to-sequence learning.

\textbf{General Audio Models.} General audio models are pretrained on broad audio corpora such as AudioSet \cite{audioset}, which include but are not limited to speech. This exposure provides them with a wider acoustic distribution than that of speech-only models. We include AudioMAE~\cite{audiomae} and WavJEPA~\cite{wavjepa}, which learn representations via self-supervised reconstruction and predictive objectives, respectively. We additionally include AST~\cite{ast}, a vision transformer adapted to audio via supervised training on AudioSet; and CLAP~\cite{clap}, trained via contrastive learning on audio-text pairs. Both of these models have been applied to audio classification tasks beyond speech~\cite{hear, ast_crying, ast_parkinsons}.

\textbf{Domain-specific Models.} Domain-specific models are pretrained on data from a targeted application domain rather than general speech or audio corpora, offering a direct test of whether specialized pretraining confers an advantage on clinical tasks. We include emotion2vec+~\cite{emotion2vec}, pretrained on emotional speech via self-supervised teacher-student distillation; and OPERA~\cite{opera}, a model pretrained on respiratory audio via a multi-task self-supervised objective.

\subsection{Evaluation Protocol}
To evaluate the performance of pretrained representations across clinical speech tasks, we adopt a linear probing protocol following standard practice \cite{opera, hear, hear_acoustic}. Encoder weights are kept frozen throughout, and a single linear layer is trained on top of the mean-pooled encoder output. The output dimensionality is set to the number of target classes for classification tasks or a single unit for regression. Compared to full fine-tuning, this approach is computationally efficient and less prone to overfitting, making it appropriate for the limited dataset sizes typical in clinical tasks.

Official data splits are used where available; otherwise, datasets are partitioned in one of three ways depending on their size. Datasets with a large number of speakers are split into training (70\%), validation (10\%), and test (20\%) sets with speaker-disjoint partitions stratified by label, sex, and age when possible. Datasets with a small number of speakers are split via 5-fold cross-validation with speaker-disjoint folds stratified by label. Finally, the official split for COVID-19 Sounds \cite{covid19sounds} is replaced with a custom speaker-disjoint partition due to speaker leakage in the original release. 

Beyond standard within-dataset evaluation, we also examine zero-shot cross-condition transfer, in which a linear probe trained on a source dataset is evaluated directly on a disjoint target dataset without any exposure to target data during training. The source dataset is split into training (80\%) and validation (20\%) sets, and the entire target dataset serves as the test set. Source and target datasets may belong to the same or different stages in Berisha and Liss' framework. We restrict zero-shot evaluation to the binary classification tasks from the main benchmark. For consistency across datasets, controls are grouped as one class and clinically affected participants as the other; for emotion datasets, non-negative emotions are mapped to controls and negative emotions to the affected class as the closest analogue to the clinical dichotomy. 
 
We evaluate classification tasks using the area under the receiver operating characteristic curve (AUC), reported as the macro-average across classes for multi-class settings. Regression tasks are evaluated using mean absolute error (MAE). For datasets evaluated on a held-out test set, 95\% confidence intervals for both AUC and MAE are estimated via bootstrap resampling ($n = 1{,}000$). For datasets evaluated under cross-validation, mean performance across folds is reported. To compare results across tasks with mixed metrics, we also report mean reciprocal rank (MRR) by ranking models on each task within a speech-production stage and averaging the reciprocal of those ranks across all tasks in the stage. Additional implementation details are provided in Appendix~\ref{app:implementation}.

\subsection{Implementation Details}
\label{subsec:compute}
Embedding extraction for all 12 encoders across the benchmark was performed on compute nodes with 8× NVIDIA H100 80GB GPUs, with extraction parallelized across multiple GPUs and taking approximately 288 GPU-hours total. GPU utilization averaged 70\% due to variable-length audio samples, with outlier samples requiring dedicated GPU memory; batching strategies could further optimize this process. Once embeddings were cached, all linear probing experiments (main benchmark, zero-shot transfer, and data efficiency) were executed locally with 8 concurrent jobs, requiring approximately 20 hours of wall-clock time.
\section{Results}

\subsection{Benchmark Evaluation}
\label{sec:results-main}
\begin{figure}[!t]
    \centering
    \includegraphics[width=0.99\linewidth]{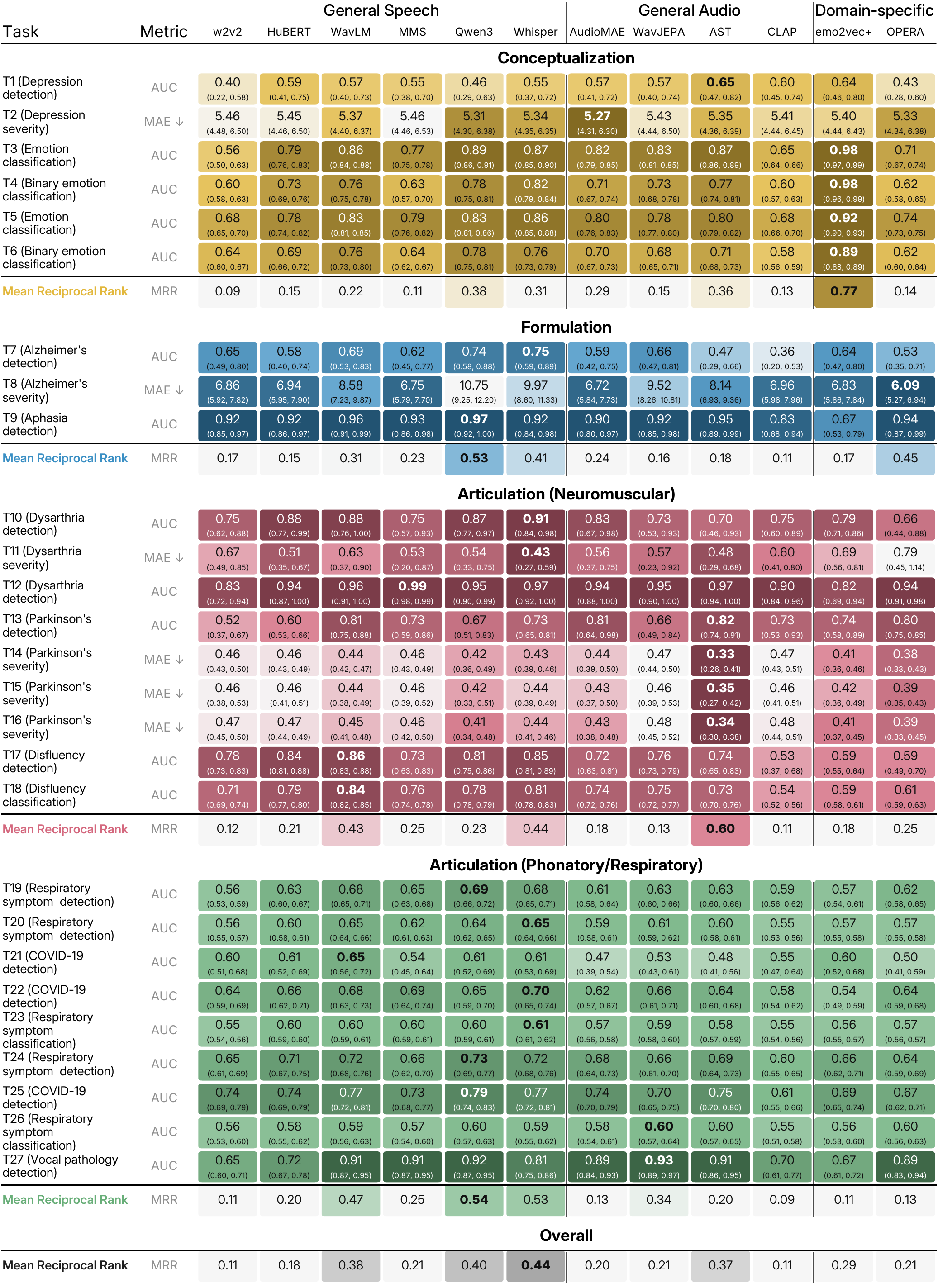}
    \caption{The benchmark evaluation of 12 audio encoders across 27 clinical speech AI tasks, grouped according to Berisha and Liss' framework of speech production~\cite{clinical_taxonomy}. Classification tasks report AUC and regression tasks report MAE, with 95\% bootstrap confidence intervals for tasks evaluated with a held-out test set or ($\mu$ - $\sigma$, $\mu$ + $\sigma$) for tasks evaluated with 5-fold cross-validation. MRR is computed within each category and overall. Model names are shortened as follows: w2v2 = wav2vec~2.0, Qwen3 = Qwen3-TTS-Tokenizer, emo2vec+ = emotion2vec+, and OPERA = OPERA-GT. 
    }
    \label{fig:main-results}
    \vspace{-2em}
\end{figure}

\autoref{fig:main-results} shows the performance of 12 state-of-the-art audio encoders across all 27 tasks in \benchmarkname according to per-task AUC and MAE results with 95\% confidence intervals, alongside category-level and overall MRR scores.
The results show that the speech-production taxonomy is a meaningful predictor of task difficulty. Conceptualization tasks are consistently difficult, with depression detection (T1) yielding an AUC between 0.40 and 0.65 and depression severity estimation (T2) showing similarly low performance across models. In contrast, tasks belonging to the formulation and neuromuscular articulation stages are more tractable. Aphasia detection (T9) reaches a maximum AUC of 0.97, and the best models exceed an AUC of 0.82 on all neuromuscular articulation classification tasks (T10, T12, T13, T17, T18). 

However, performance is not determined by the speech-production stage alone. Although respiratory tasks directly affect breath support and vocal function, performance remains modest across models, with the best AUC reaching 0.79 for COVID-19 detection (T25). In contrast, vocal pathology detection (T27) is substantially more separable, with WavJEPA achieving an AUC of 0.93. This contrast suggests that heterogeneous recording conditions and label noise can attenuate the expected relationship between production-stage proximity and predictive performance. Unlike other datasets, several in the respiratory category are crowdsourced across diverse devices and environments, introducing acquisition variability and confounding factors that obscure the clinical signal.

Overall, no single encoder achieves uniformly strong performance across all conditions impacting speech. The strongest overall models are Whisper (MRR: 0.44), Qwen3-TTS-Tokenizer (MRR: 0.40), and WavLM (MRR: 0.38), whereas CLAP and wav2vec~2.0 are among the weakest performers. The strong aggregate performance of Whisper and Qwen3 suggests an association between scale and broader clinical utility, reflecting their status as the models with the most extensive pretraining. However, overall MRR masks stage-specific variation; emotion2vec+ dominates conceptualization tasks (MRR: 0.77), AST and Whisper perform best on neuromuscular articulation tasks (MRR: 0.60 and 0.44), and Qwen3 and Whisper lead phonatory / respiratory tasks (MRR: 0.54 and 0.53). These discrepancies indicate that current models encode different dimensions of clinical speech variation, but none provides a representation that generalizes reliably across clinical domains.

Domain-specific pretraining yields selective gains. The strong performance of emotion2vec+ on emotion tasks (T3--T6) is likely due to the inclusion of the IEMOCAP dataset in its pretraining corpus~\cite{emotion2vec}. Its poor performance on depression detection, a closely related conceptualization-stage condition, illustrates that domain-specific pretraining does not generalize to even clinically proximate tasks. OPERA, another domain-specific model, shows a similar mismatch. Although pretrained on respiratory audio, it performs weakly on respiratory tasks (MRR: 0.13) and is instead most competitive on formulation tasks (MRR: 0.45). These patterns suggest that narrow domain alignment between pretraining and downstream tasks is insufficient for general clinical speech understanding. Finally, to assess performance under data scarcity, a central constraint in clinical settings, we conduct data efficiency experiments varying training set size from 12.5\% to 100\% for the top three models, with results reported in Appendix~\ref{app:data_efficiency}.

\subsection{Zero-shot Cross-condition Transfer}
\label{sec:zero-shot}
\begin{figure}[!t]
    \centering
    \includegraphics[width=0.95\linewidth]{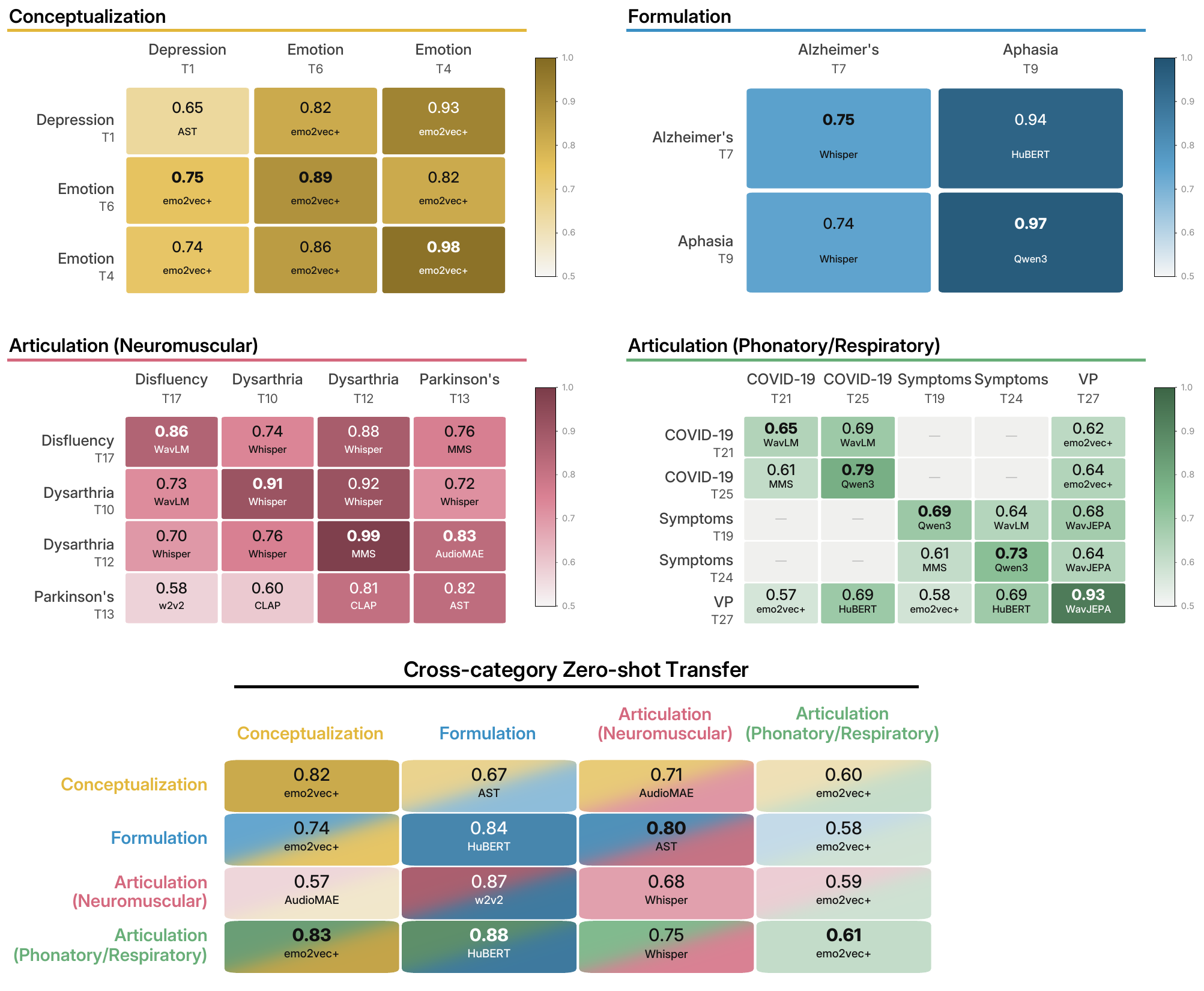}
    \caption{The evaluation of zero-shot transfer for classification tasks. The top four grids show transfer across tasks within the same category, while the bottom grid shows averaged cross-category performance. Each cell reports the best model and the AUC it achieved for the given task.}
    \label{fig:zero-shot}
    \vspace{-2em}
\end{figure}

To assess whether the representations learned by current audio encoders capture clinical structure beyond dataset-specific artefacts, we evaluate zero-shot transfer between datasets. 
\autoref{fig:zero-shot} reports the best zero-shot AUC and the corresponding model for each source--target pair; for cross-category transfer, source and target datasets are pooled within each stage. Complete results for all models and all transfer pairs are provided in Appendix~\ref{app:zero-shot}.

\paragraph{Within-category transfer.}
For many source--target pairs within a production stage, the best zero-shot transfer approaches within-dataset performance, with several cases meeting or exceeding it. A probe trained on DementiaBank reaches an AUC of 0.94 on aphasia detection (HuBERT) compared to a within-dataset best of 0.97, and probes trained on either emotion dataset reach AUCs of 0.74--0.75 on depression detection (emotion2vec+) compared to a within-dataset best of 0.65. Transfer is consistently weaker across phonatory / respiratory tasks (AUC: 0.57--0.69), identifying this category as the most resistant to cross-dataset generalization. Notably, the best zero-shot model differs from the best within-dataset model in approximately half of all source--target pairs. For example, AudioMAE exceeds the best within-dataset performance on Parkinson's detection when trained on UASpeech, and Whisper leads dysarthria-to-dysarthria transfer over the within-dataset leader MMS, indicating that within-dataset and zero-shot regimes reward different representational properties.
 
\paragraph{Cross-condition transfer.}
Across production stages, transfer is variable but reveals a clear asymmetry. Representations trained on phonatory / respiratory data transfer into conceptualization and formulation tasks at AUCs of 0.83 (emotion2vec+) and 0.88 (HuBERT) respectively, while transfer in the reverse direction does not exceed an AUC of 0.60. Formulation also transfers effectively into neuromuscular tasks (AUC: 0.80, AST). Notably, model rankings observed in the within-dataset evaluation do not predict cross-category transfer performance. Qwen3 achieves the highest within-dataset performance on phonatory / respiratory tasks but fails to transfer across categories, while emotion2vec+, which shows mid-tier within-dataset performance leads cross-category transfer to phonatory / respiratory targets. Progress on clinical speech AI will require representations that maintain performance across regimes, rather than models that excel within a single dataset; closing this gap remains an open challenge for future work. 

\section{Broader Impact and Limitations}
\label{sec:discussion}
Speech-based health assessment could transform everyday devices into diagnostic tools, extending healthcare access to underserved populations. The ubiquity of microphones available in smartphones, smart speakers, and even earbuds enables frequent monitoring with minimal user effort. These opportunities not only circumvent the logistical barriers inherent in traditional clinical assessments but also enable the capture of subtle physiological shifts.

Realizing the full diagnostic potential of clinical speech AI requires representations that generalize across conditions, populations, and recording settings. In pursuit of this objective, \benchmarkname provides a unified evaluation infrastructure to drive progress toward clinically deployable models while ensuring that performance gains are meaningful rather than driven by dataset idiosyncrasies.

One key consideration for \benchmarkname is the utility and objectivity of the underlying labels for each dataset.
The majority of tasks are designed to reveal acoustic markers that classify people with and without a medical condition.
The resulting markers would ideally be used to screen people with nascent symptoms; however, these datasets typically comprise people who are either completely healthy or have lived with the condition for multiple years.
These datasets also represent single assessments rather than disease progression, though many conditions vary significantly over time.
Several other tasks rely on patients' or clinicians' answers to standardized questionnaires (e.g., PHQ-8~\cite{phq8}, UPDRS~\cite{updrs}).
Although the questionnaires are clinically validated and human-reported responses should not be discredited, there is unavoidable subjectivity associated with anchoring bias, recall bias, and person-specific internal scaling that may obscure physiological information~\cite{recall_bias, selfreport_bias}.

Another consideration relates to the diversity of human subjects across the benchmark.
\benchmarkname primarily comprises English-language recordings, though diverse regional accents are represented through datasets collected outside North America. 
We did not conduct a formal analysis of how this factor might have confounded our zero-shot analysis, so this remains an opportunity for future work.
Sex and age are other demographic factors that warrant further investigation, particularly since certain populations are predisposed to specific conditions (e.g., elderly women and dementia~\cite{women_dementia}).

Lastly, \benchmarkname does not exhaustively cover the clinical domains in which speech-based assessment has shown promise. Conditions such as Huntington's disease~\cite{hertrich1994acoustic}, pediatric speech sound disorders~\cite{dodd2014differential}, and chronic obstructive pulmonary disease~\cite{pulmolistener, copd_features} are a subset of the diverse clinical domains where vocal biomarkers have shown promise, yet publicly available datasets are either unavailable or not sizable enough for thorough evaluation.
Subsequent iterations will incorporate these and other emerging areas to provide an increasingly comprehensive evaluation of clinical speech AI.
\section{Conclusion}
\benchmarkname establishes a comprehensive evaluation framework for clinical speech AI. 
While large-scale general-purpose speech models currently provide the most robust baselines, experiments with our benchmark reveal that no single representation yet generalizes reliably across the diverse clinical landscape. 
Moreover, cross-condition transfer analyses demonstrate critical gaps in model robustness and data efficiency.
Ultimately, \benchmarkname serves as a foundation for tracking progress toward truly generalizable representations capable of supporting real-world clinical monitoring.




\bibliographystyle{unsrt}
\bibliography{references}





\newpage
\appendix
\section*{Appendix for \benchmarkname}
\begin{enumerate}[label=\textbf{\Alph*}, leftmargin=3em]
    \item \hyperref[app:datasets]{\textbf{Dataset Details}} \dotfill \pageref{app:datasets}
    \item \hyperref[app:tasks]{\textbf{Task Details}} \dotfill \pageref{app:tasks}
    \item \hyperref[app:models]{\textbf{Model Details}} \dotfill \pageref{app:models}
    \item \hyperref[app:implementation]{\textbf{Implementation Details}} \dotfill \pageref{app:implementation}
    \item \hyperref[app:data_efficiency]{\textbf{Data Efficiency Analysis}} \dotfill \pageref{app:data_efficiency}
    \item \hyperref[app:zero-shot]{\textbf{Zero-shot Transfer Analysis}} \dotfill \pageref{app:zero-shot}
\end{enumerate}
\vspace{2em}
\section{Dataset Details}
\label{app:datasets}

This section provides a comprehensive description of the datasets utilized within the \benchmarkname benchmark. The datasets are categorized according to the specific stage of speech production they represent. For each dataset, we describe the recruitment procedure, study protocol, and labeling process. Their access methods and licenses are listed in \autoref{tab:dataset-access}.

\subsection{Conceptualization}
\textbf{Extended Distress Analysis Interview Corpus, Wizard-of-Oz (EDAIC-WOZ)}~\cite{edaic} was developed at the Institute for Creative Technologies at the University of Southern California. A total of 275 participants (170~male, 105~female) drawn from the general population and American military veterans completed semi-structured clinical interviews with a virtually animated interviewer covering topics related to psychological distress, daily functioning, and mood, lasting approximately 16~minutes. For the 163~training and 56~validation participants, the interviewer was controlled by a human operator via a Wizard-of-Oz paradigm, while the 56~test participants were interviewed by a fully autonomous system, introducing a distributional shift relative to the training data. Prior to each interview, participants completed the Patient Health Questionnaire (PHQ-8) for depression severity~\cite{phq8}; scores of 10 or above indicate a depressive disorder, yielding binary depression labels. 

\textbf{Ryerson Audio-Visual Database of Emotional Speech and Song (RAVDESS)}~\cite{ravdess} comprises 24~professional actors (12~female, 12~male) narrating two lexically matched statements (\textit{``Kids are talking by the door''} and \textit{``Dogs are sitting by the door''}) in a neutral North American English accent while expressing one of eight emotions: neutral, calm, happy, sad, angry, fearful, disgust, and surprised. Each emotion (except neutral) was produced at two intensity levels (normal and strong), yielding 60~recordings per actor. The speech subset is sampled at 48~kHz. Each recording was rated 10~times for emotional validity, intensity, and authenticity by 247~untrained participants with high inter-rater reliability. The dataset is fully balanced across actors, emotions, and intensity levels.
 
\textbf{Interactive Emotional Dyadic Motion Capture (IEMOCAP)}~\cite{iemocap} was collected at the Speech Analysis and Interpretation Laboratory at the University of Southern California. The corpus contains approximately 12~hours of audiovisual recordings from 10~actors (5~male, 5~female) across five dyadic sessions, each with a unique male-female actor pair. Within each session, actors performed both scripted emotional dialogues and improvised hypothetical scenarios designed to elicit a range of emotional states. Utterances were segmented and annotated by trained raters using categorical emotion labels and dimensional ratings (valence, activation, dominance). The original annotations include neutral, happiness, sadness, anger, surprise, fear, disgust, frustration, excited, and other emotional states. Following standard practice~\cite{iemocap}, we use a four-class subset in which \textit{excited} is merged with \textit{happy}, \textit{frustration} with \textit{anger}, and \textit{fear}, \textit{disgust}, \textit{surprise}, and \textit{other} are excluded due to their limited representation.

\begin{table}[tbp]
\centering
\scriptsize
\caption{Access and licensing information for each dataset in \benchmarkname.}
\label{tab:dataset-access}
\resizebox{\textwidth}{!}{%
\begin{tabular}{>{\raggedright\arraybackslash}p{0.11\linewidth} >{\raggedright\arraybackslash}p{0.10\linewidth} >{\raggedright\arraybackslash}p{0.49\linewidth} >{\raggedright\arraybackslash}p{0.10\linewidth} >{\raggedright\arraybackslash}p{0.15\linewidth}}
\toprule
\textbf{Category} & \textbf{Dataset} & \textbf{Link} & \textbf{Availability} & \textbf{License} \\
\midrule
Conceptualization & EDAIC-WOZ \cite{edaic} & \url{https://dcapswoz.ict.usc.edu/} & Available on request & DAIC-WOZ EULA \\
 & RAVDESS \cite{ravdess} & \url{https://zenodo.org/records/1188976} & Open access & CC BY-NC-SA 4.0 \\
 & IEMOCAP \cite{iemocap} & \url{https://sail.usc.edu/iemocap/iemocap_release.htm} & Available on request & Custom \\
\midrule
Formulation & DementiaBank \cite{dementiabank, address-M} &   \url{https://talkbank.org/dementia/} & Available on request & TalkBank CC BY-NC-SA 3.0  \\

 & AphasiaBank \cite{aphasiabank} & \url{https://talkbank.org/aphasia/} & Available on request & TalkBank CC BY-NC-SA 3.0\\
\midrule
Articulation (Neuromuscular) & TORGO \cite{torgo} & \url{https://www.cs.toronto.edu/~complingweb/data/TORGO/torgo.html} & Open access & Custom \\
 & UASpeech \cite{uaspeech} & \url{https://speechtechnology.web.illinois.edu/uaspeech/} & Available on request & Custom \\
 & MDVR-KCL \cite{mdvrkcl} & \url{https://zenodo.org/records/2867216} & Open access & CC BY 4.0 \\
 & KSoF-C \cite{ksofc_challenge} & \url{https://zenodo.org/records/7258757} & Available on request & EULA KSoF Challenge \\
\midrule
Articulation (Phonatory / Respiratory) & COVID-19 Sounds \cite{covid19sounds} & \url{https://covid-19-sounds.org/en/blog/neurips_dataset} & Available on request & Custom \\
 & Coswara \cite{coswara} & \url{https://github.com/iiscleap/Coswara-Data} & Open access & CC BY 4.0 \\
 & AVFAD \cite{avfad} & \url{https://acsa.web.ua.pt/AVFAD.htm} & Available on request & Custom \\
\bottomrule
\end{tabular}
}
\end{table}

\subsection{Formulation}
\textbf{DementiaBank} (ADReSS-M subset~\cite{address-M}) was collected as part of the Alzheimer and Related Dementias Study at the University of Pittsburgh. Participants completed the Cookie Theft Picture Description Task from the Boston Diagnostic Aphasia Examination~\cite{bdae}, producing spontaneous speech descriptions of a kitchen scene. Participants also completed the Mini-Mental State Examination (MMSE)~\cite{mmse} to assess their level of cognitive impairment. We use the ADReSS-M challenge subset~\cite{address-M}, released for the ICASSP 2023 Signal Processing Grand Challenge on multilingual Alzheimer's dementia recognition. The subset is statistically matched to mitigate common evaluation biases including repeated recordings from the same participant, audio quality variation, and demographic imbalance. The training partition consists of English-language recordings, and the test partition consists of Greek-language recordings from a separate cohort, making this the only cross-lingual evaluation in the benchmark. 

\textbf{AphasiaBank}~\cite{aphasiabank} is a publicly available database of structured discourse samples from individuals with post-stroke aphasia and neurologically healthy controls, hosted by the TalkBank consortium. We use a subset drawn from five sites --- Adler, Kansas, Kurland, SCALE, and Wright --- retaining first-session recordings only, yielding 206 participants (119 persons with aphasia, 87 controls). Persons with aphasia ranged in age from 36 to 91 years and spanned mild to severe impairment as measured by the Western Aphasia Battery Aphasia Quotient~\cite{wab} (mean 67.6~$\pm$~22.1). Aphasia types included Anomic, Broca's, Conduction, Wernicke's, and several transcortical variants. Each participant completed the AphasiaBank Discourse Protocol, from which we extracted four discourse tasks: (1) Cinderella story retell, (2) window picture sequence, (3) umbrella picture sequence, and (4) cat picture sequence. 

\subsection{Articulation (Neuromuscular)}
\textbf{TORGO}~\cite{torgo} contains approximately 8~hours of English speech from 15 speakers: 8 with dysarthria (5~male, 3~female) resulting from cerebral palsy or ALS, and 7 age- and gender-matched healthy controls (4~male, 3~female). Speech was recorded using a head-mounted microphone and a directional microphone, but we only use the head-mounted channel. Speech material was drawn from multiple sources including the TIMIT database \cite{timit}, lists of identified phonetic contrasts, and standardized speech intelligibility assessments. Each dysarthric speaker was assessed by a speech-language pathologist using the Frenchay Dysarthria Assessment~\cite{frenchay}. 
 
\textbf{UASpeech}~\cite{uaspeech} contains isolated word recordings from 15~speakers with cerebral palsy and 13~age-matched healthy controls. Each participant uttered 765~isolated words: 300~uncommon words, 3~repetitions of 100~common words, and 165~digits, radio alphabet letters, and computer commands. Audio was recorded using an 8-microphone array, but we use the single close-talking microphone channel. Intelligibility was assessed by five listeners who transcribed the recordings word by word; each speaker's intelligibility score was computed as the average percentage of words correctly transcribed. Speakers were subsequently grouped into four intelligibility categories: very low, low, mid, and high. The distribution of dysarthric speakers across these categories is skewed toward lower intelligibility, with the very low and low groups together accounting for the majority.
 
\textbf{MDVR-KCL}~\cite{mdvrkcl} was released by King's College London and contains speech recordings from 37~participants: 16 with Parkinson's disease and 21~healthy controls. Recordings were made at 44.1\,kHz in a phone-call scenario where participants held the recording device to their preferred ear. Participants read two passages: the fable \textit{North Wind and the Sun} and an excerpt from \textit{Computer Applications in Geography}. Each participant was assessed along three clinically validated scales by experts: the Hoehn and Yahr (H\&Y) scale~\cite{hoehnyahr}, Unified Parkinson’s Disease Rating Scale (UPDRS) Part~II Item~5 (activities of daily living and speech), and UPDRS Part~III Item~18 (motor examination and speech) \cite{updrs}. 

\textbf{KSoF-C}~\cite{ksofc_challenge} is a subset of the Kassel State of Fluency (KSoF) dataset~\cite{ksof}, a German-language stuttering corpus recorded during therapy sessions. The full KSoF corpus contains 5,597~three-second clips extracted from 214~recordings of 37~persons who stutter (28~male, 9~female). A distinctive property of this corpus is that all speakers had undergone fluency-shaping therapy, making it the only public dataset to include post-therapy stuttered speech; therefore, speakers exhibit both dysfluent events and therapy-specific speech modifications alongside fluent speech. Clips were annotated with one of six labels by three trained annotators: blocks, prolongations, sound repetitions, word repetitions, interjections, and speech modifications. KSoF-C is the subset released for the ACM Multimedia 2022 ComParE challenge \cite{ksofc_challenge}; we use this split with its predefined train, validation, and test partitions. 

\subsection{Articulation (Phonatory / Respiratory)}
\textbf{COVID-19 Sounds} \cite{covid19sounds} was collected by the University of Cambridge via a smartphone and web application. We use the speech recordings, in which participants read the sentence \textit{``I hope my data can help to manage the virus pandemic''} three times in their native language. The full corpus contains 53,449~samples from 36,116~participants, totaling over 552~hours of audio. Along with audio, the participants self-reported their COVID-19 status and respiratory symptoms. The dataset comes with two curated subsets widely used for benchmarking: one balanced for respiratory symptoms and the other balanced for COVID-19 status, each screened for recording quality and released as controlled English-language partitions. 
 
\textbf{Coswara} \cite{coswara} was collected by the Indian Institute of Science via Android and Web applications. We use the normal-pace counting speech recordings, in which participants count from 1 to 20 in English. Participants self-reported their symptoms and COVID-19 status by selecting one of three categories: negative, positive, or recovered. The control group is heterogeneous, comprising completely healthy individuals alongside those with respiratory ailments and COVID-like symptoms, reflecting realistic variability in population-level screening settings.
 
\textbf{Advanced Voice Function Assessment Databases (AVFAD)}~\cite{avfad} contains Portuguese recordings from 709~individuals: 346~with clinically diagnosed vocal pathology and 363~without vocal alterations. The pathological group encompasses 26~distinct diagnoses, the most prevalent being vocal fold nodules, polyps, cysts, and Reinke's edema. All diagnoses were registered according to the Classification Manual of Voice Disorders-I~\cite{cmvd}. 

\section{Task Details}
\label{app:tasks}

This section provides a detailed description of all 27 tasks formulated in the \benchmarkname benchmark. The tasks are again organized by the stage of speech production. For each task, we describe the prediction target, the dataset used, and any relevant details regarding label construction or subset selection. The demographic statistics are summarized in \autoref{tab:task-details}. 

\subsection{Conceptualization}
\textbf{T1.} This task aims to classify speakers in the EDAIC-WOZ dataset \cite{edaic} as either depressed or healthy. The EDAIC-WOZ dataset contains clinical interview recordings of participants scored on the PHQ-8 depression scale~\cite{phq8}, where speakers with a score of 10 or above are labeled as depressed.

\clearpage
\begingroup
\setlength{\tabcolsep}{3pt}
\hyphenpenalty=200
\exhyphenpenalty=200
\scriptsize
\begin{longtable}{>{\raggedright\arraybackslash}p{0.15\linewidth} >{\raggedright\arraybackslash}p{0.10\linewidth} >{\raggedright\arraybackslash}p{0.025\linewidth} >{\raggedright\arraybackslash}p{0.075\linewidth} >{\raggedright\arraybackslash}p{0.09\linewidth} >{\raggedright\arraybackslash}p{0.09\linewidth} >{\raggedright\arraybackslash}p{0.32\linewidth}}
\caption{Demographic data and label distributions for each task in \benchmarkname. Label distributions are reported as healthy \% / disease \% for binary classification tasks and as [range of label], mean~$\pm$~standard deviation for regression tasks. Details are directly reported from publications and documentation.}\label{tab:task-details}\\
\toprule
\textbf{Category} & \textbf{Dataset} & \textbf{ID} & \textbf{Country} & \textbf{Age} & \textbf{Sex} & \textbf{Label Distribution} \\
\midrule
\endfirsthead
\multicolumn{7}{l}{\textit{(Table~\ref{tab:task-details} continued)}} \\
\toprule
\textbf{Category} & \textbf{Dataset} & \textbf{ID} & \textbf{Country} & \textbf{Age} & \textbf{Sex} & \textbf{Label Distribution} \\
\midrule
\endhead
\midrule
\multicolumn{7}{r}{\textit{Continued on next page}} \\
\endfoot
\bottomrule
\endlastfoot
  Conceptual\-ization & EDAIC-WOZ~\cite{edaic} & T1 & USA & N/A & M: 61.8\%\newline F: 38.2\% & 24.0\% / 76.0\% \\
   &  & T2 & USA & N/A & M: 61.8\%\newline F: 38.2\% & $[0, 23],\ \mu = 6.94 \pm 6.11$ \\
\cmidrule(lr){2-7}
 & RAVDESS~\cite{ravdess} & T3 & Canada & N/A & M: 50\%\newline F: 50\% & Calm: 13.3\%; Happy: 13.3\%; Sad: 13.3\%; Angry: 13.3\%; Fearful: 13.3\%; Disgust: 13.3\%; Surprised: 13.3\%; Neutral: 6.7\% \\
 &  & T4 & Canada & N/A & M: 50\%\newline F: 50\% & 53.3\% / 46.7\% \\
\cmidrule(lr){2-7}
 & IEMOCAP~\cite{iemocap} & T5 & USA & N/A & M: 50.8\%\newline F: 49.2\% & Neutral: 23.1\%; Anger / Frustrated: 40.0\%; Sad: 14.7\%; Happy / Excited: 22.2\% \\
 &  & T6 & USA & N/A & M: 50.8\%\newline F: 49.2\% & 54.7\% / 45.3\% \\
\midrule
Formulation & Dementia\-Bank~\cite{dementiabank} & T7 & USA; Greece & $\mu = 68.1 \pm 7.0$ & M: 33.0\%\newline F: 67.0\% & 50.9\% / 49.1\% \\
 &  & T8 & USA; Greece & $\mu = 68.0 \pm 7.0$ & M: 32.8\%\newline F: 67.2\% & $[3, 30],\ \mu = 23.58 \pm 6.61$ \\
\cmidrule(lr){2-7}
 & Aphasia\-Bank~\cite{aphasiabank} & T9 & USA & $\mu = 62.4 \pm 13.6$ & M: 55.4\%\newline F: 44.2\% & 72.0\% / 28.0\% \\
\midrule
Articulation (neuro\-muscular) & TORGO~\cite{torgo} & T10 & Canada & N/A & M: 58.1\%\newline F: 41.9\% & 33.8\% / 66.2\% \\
 &  & T11 & Canada & N/A & M: 55.0\%\newline F: 45.0\% & $[1, 3],\ \mu = 1.84 \pm 0.94$ \\
\cmidrule(lr){2-7}
 & UASpeech~\cite{uaspeech} & T12 & USA & N/A & M: 71.4\%\newline F: 28.6\% & 53.6\% / 46.4\% \\
\cmidrule(lr){2-7}
 & MDVR-KCL~\cite{mdvrkcl} & T13 & UK & N/A & N/A & 42.5\% / 57.5\% \\
 &  & T14 & UK & N/A & N/A & $[0, 3],\ \mu = 0.34 \pm 0.65$ \\
 &  & T15 & UK & N/A & N/A & $[0, 3],\ \mu = 0.40 \pm 0.70$ \\
 &  & T16 & UK & N/A & N/A & $[0, 4],\ \mu = 1.10 \pm 1.36$ \\
\cmidrule(lr){2-7}
 & KSoF-C~\cite{ksof} & T17 & Germany & N/A & M: 86.0\%\newline F: 14.0\% & 74.1\% / 25.9\% \\
 &  & T18 & Germany & N/A & M: 86.2\%\newline F:~13.8\% & Block:~20.7\%; Prolongation:~12.0\%; Sound~rep.:~14.8\%; Word rep.:~3.9\%; Modified:~24.4\%; Interjection:~13.0\%; No dysfluency:~24.7\%; Garbage:~3.1\% \\
\midrule
Articulation (phonatory / respiratory) & COVID-19 Sounds~\cite{covid19sounds} & T19 & Intl. & $\mu = 39.4 \pm 14.1$ & M: 46.4\%\newline F: 52.3\%\newline Unk.: 1.2\% & 48.7\% / 51.3\% \\
 &  & T20 & Intl. & $\mu = 39.2 \pm 15.2$ & M: 61.6\%\newline F: 37.2\%\newline Unk.: 1.2\% & 52.5\% / 47.5\% \\
 &  & T21 & Intl. & $\mu = 40.0 \pm 13.7$ & M: 50.9\%\newline F: 48.4\%\newline Unk.: 0.7\% & 49.4\% / 50.6\% \\
 &  & T22 & Intl. & $\mu = 41.3 \pm 14.4$ & M: 56.0\%\newline F: 42.9\%\newline Unk.: 1.1\% & 19.6\% / 80.4\% \\
 &  & T23 & Intl. & $\mu = 36.9 \pm 14.1$ & M: 56.3\%\newline F: 42.6\%\newline Unk.: 1.2\% & Dry cough:~49.2\%; Wet cough:~23.0\%; Fever:~8.1\%; Headache:~22.9\%; Muscle ache:~14.2\%; Dizziness:~6.7\%; Sore throat:~27.8\%; Short breath:~13.0\%; Tightness:~12.8\%; Runny/blocked nose:~5.3\%; Smell/taste loss:~6.1\%; Runny:~16.3\% \\
\cmidrule(lr){2-7}
 & Coswara~\cite{coswara} & T24 & India & $\mu = 35.1 \pm 14.1$ & M: 69.0\%\newline F: 30.9\%\newline Unk.: 0.1\% & 38.6\% / 61.4\% \\
 &  & T25 & India & $\mu = 35.2 \pm 13.9$ & M: 69.9\%\newline F: 30.0\%\newline Unk.: 0.1\% & 32.5\% / 67.5\% \\
 &  & T26 & India & $\mu = 38.6 \pm 16.0$ & M: 62.7\%\newline F: 37.2\%\newline Unk.: 0.1\% & Cold:~46.4\%; Cough:~61.9\%; Fever:~38.7\%; Diarrhoea:~5.0\%; Loss of smell:~16.2\%; Muscular pain:~30.9\%; Breathing difficulty:~20.1\%; Fatigue:~36.3\%; Sore throat:~28.0\%; Other respiratory:~6.8\% \\
\cmidrule(lr){2-7}
 & AVFAD~\cite{avfad} & T27 & Portugal & $\mu = 52.3 \pm 15.8$ & M: 29.6\%\newline F: 70.4\% & 48.8\% / 51.2\% \\
\end{longtable}
\endgroup

\textbf{T2.} This task aims to regress the continuous PHQ-8 depression severity score for each speaker in the EDAIC-WOZ dataset \cite{edaic}, using the same recordings as in T1.

\textbf{T3.} This task aims to classify speakers in the RAVDESS dataset \cite{ravdess} into one of seven emotion categories: neutral, calm, happy, sad, angry, fearful, disgust, or surprised. The RAVDESS dataset contains acted speech recordings from 24 professional actors, with each emotion represented equally across speakers.

\textbf{T4.} This task aims to classify speakers in the RAVDESS dataset \cite{ravdess} as either expressing negative or non-negative emotion using the same recordings as T3. Negative emotions comprise sad, angry, fearful, and disgust. Non-negative emotions comprise neutral, calm, happy, and surprised.

\textbf{T5.} This task aims to classify recordings in the IEMOCAP dataset \cite{iemocap} into one of four emotion categories: angry, neutral, happy, or sad. The IEMOCAP dataset contains dyadic conversation recordings across five sessions, with emotion labels assigned through majority agreement among multiple annotators.

\textbf{T6.} This task aims to classify speakers in the IEMOCAP dataset \cite{iemocap} as either expressing negative or non-negative emotion using the same recordings as T5. Negative emotions comprise angry and sad. Non-negative emotions comprise neutral and happy.

\subsection{Formulation}

\textbf{T7.} This task aims to classify speakers in the ADReSS-M challenge \cite{address-M} as either having Alzheimer's disease or being cognitively healthy. The ADReSS-M challenge is a subset of the DementiaBank dataset \cite{dementiabank} containing picture description recordings from the Cookie Theft task, with binary labels derived from clinical diagnosis of Alzheimer's disease.

\textbf{T8.} This task aims to regress the continuous MMSE cognitive severity score for each speaker in the DementiaBank dataset, using the same picture description recordings as T7. MMSE scores range from 0 to 30, with lower scores indicating greater cognitive impairment.

\textbf{T9.} This task aims to classify speakers in the AphasiaBank dataset \cite{aphasiabank} as either aphasic or healthy. The AphasiaBank dataset contains standardized discourse recordings from speakers of varying etiology and severity alongside healthy controls.

\subsection{Articulation (Neuromuscular)}

\textbf{T10.} This task aims to classify speakers in the TORGO dataset \cite{torgo} as either dysarthric or healthy. The TORGO dataset contains speech recordings from speakers with motor speech disorders of varying etiology and severity alongside healthy controls.

\textbf{T11.} This task aims to regress the dysarthria severity score for each speaker in the TORGO dataset \cite{torgo}, using the same recordings as T10. Severity is rated on a scale from 0 to 3, where higher scores indicate greater articulatory impairment.

\textbf{T12.} This task aims to classify speakers in the UASpeech dataset \cite{uaspeech} as either dysarthric or healthy. The UASpeech dataset contains speech recordings from speakers with cerebral palsy alongside healthy controls, with dysarthria severity ranging from mild to profound.

\textbf{T13.} This task aims to classify speakers in the MDVR-KCL dataset \cite{mdvrkcl} as either having Parkinson's disease or being healthy. The MDVR-KCL dataset contains read speech and spontaneous monologue recordings from speakers with Parkinson's disease alongside healthy controls.

\textbf{T14.} This task aims to regress the UPDRS Part~II Item~5 score for each speaker in the MDVR-KCL dataset \cite{mdvrkcl}, using the same recordings as T13. UPDRS Part~II Item~5 is a patient-reported measure of speech impairment in daily living, scored from 0 to 3.

\textbf{T15.} This task aims to regress the UPDRS Part~III Item~18 score for each speaker in the MDVR-KCL dataset, using the same recordings as T13. UPDRS Part~III Item~18 is a clinician-rated measure of speech quality during motor examination, scored on a scale from 0 to 3.

\textbf{T16.} This task aims to regress the H\&Y scale score for each speaker in the MDVR-KCL dataset, using the same recordings as T13. The Hoehn and Yahr scale is a clinician-rated measure of Parkinson's disease progression scored from 0 to 4, where higher scores indicate greater motor impairment.

\textbf{T17.} This task aims to classify speakers in the KSoF-C dataset \cite{ksof} as either disfluent or fluent. The KSoF-C dataset contains read speech recordings from speakers who stutter alongside fluent controls, with disfluency labels assigned at the utterance level.

\textbf{T18.} This task aims to classify the disfluency type for each utterance in the KSoF-C dataset. This is a multi-label classification task, with each utterance being assigned one or more of six disfluency categories: modified speech, blocks, sound repetitions, interjections, prolongations, and word repetitions. Two additional classes are included to account for utterances with no disfluency and those with poor audio quality or recording errors.

\subsection{Articulation (Phonatory / Respiratory)}
\textbf{T19.} This task aims to classify speakers in the COVID-19 Sounds dataset \cite{covid19sounds} as either symptomatic or healthy. The dataset contains crowdsourced speech recordings from participants who reported their symptoms at the time of participation. Those in the symptomatic group reported at least one of the following respiratory symptoms: dry cough, sore throat, wet cough, headache, muscle ache, shortness of breath, chest tightness, fever, dizziness, smell/taste loss or runny nose. This task uses an official curated subset with controlled recording quality, balanced class distribution, and primarily English recordings.

\textbf{T20.} The dataset and prediction target are the same as T19, but T20 is based on the full COVID-19 Sounds dataset \cite{covid19sounds} rather than the curated subset, resulting in a larger sample size with greater variation in recording quality, language, and class balance.

\textbf{T21.} This task aims to classify speakers in the COVID-19 Sounds dataset \cite{covid19sounds} as either COVID-19 positive or negative based on their PCR test result. This task uses an official curated subset specifically balanced for COVID-19 status, with controlled recording quality and primarily English recordings.

\textbf{T22.} The dataset and prediction target are the same as T21, but T22 is based on the full COVID-19 Sounds dataset \cite{covid19sounds} rather than the curated subset, resulting in a heavily imbalanced class distribution with 20\% positive and 80\% negative cases.

\textbf{T23.} This task aims to classify the respiratory symptom type for each speaker in the COVID-19 Sounds dataset \cite{covid19sounds}. As a multi-label classification task, only speakers who reported at least one symptom are included, with each speaker labeled with one or more of the following symptom categories: dry cough, sore throat, wet cough, headache, muscle ache, shortness of breath, chest tightness, fever, dizziness, smell/taste loss, runny or blocked nose.

\textbf{T24.} This task aims to classify speakers in the Coswara dataset \cite{coswara} as either symptomatic or healthy, where the symptomatic group consists of participants who reported at least one of the following respiratory symptoms: cough, cold, fever, fatigue, muscle pain, sore throat, breathing difficulty, loss of smell, and diarrhoea.

\textbf{T25.} This task aims to classify speakers in the Coswara dataset \cite{coswara} as either COVID-19 positive or negative, using the same recordings as T24. The COVID-19 labels are based on self-reported test results and only speakers who reported a test outcome are included.

\textbf{T26.} This task aims to classify the respiratory symptom type for each speaker in the Coswara dataset \cite{coswara}. As a multi-label classification task, only speakers who reported at least one symptom are included, with each speaker labeled with one or more of the following symptom categories: cough, cold, fever, fatigue, muscle pain, sore throat, breathing difficulty, loss of smell, and diarrhoea. 

\textbf{T27.} This task aims to classify speakers in the AVFAD dataset \cite{avfad} as either having a vocal pathology or not. The AVFAD dataset contains multiple speech tasks, and we use the read speech recordings for this task.

\section{Model Details}
\label{app:models}
This section describes the models used in our benchmark, including their training objectives, pretraining data, and the specific variants used. All models are used with publicly available pretrained weights without additional fine-tuning unless otherwise specified.

\begin{table*}[t]
\centering

\caption{The models evaluated on \benchmarkname. Pretraining dataset sizes are approximated for models trained on aggregated or pseudo-labeled datasets. FT indicates additional fine-tuning on labeled data.}
\label{tab:model_training_data}
\scriptsize
\begin{tblr}{
  width = \textwidth,
  colspec = {
    Q[2.0,l]
    Q[2.3,l]
    Q[1.2,l]
    Q[2.0,l]
    Q[2.0,l]
    Q[2.0,c]
    Q[0.8,c]
  },
  row{1} = {font=\bfseries},
  rowsep = 2pt,
  colsep = 4pt
}
\toprule
Model & Objective & Input & Supervision & Pretraining Source & Pretraining Amount (Hours) & Params (M) \\
\midrule
wav2vec~2.0 \cite{wav2vec2} & Contrastive masked prediction & Raw audio & Self-supervised + supervised FT & Libri-Light; FT on~LibriSpeech & 60k + 960 & 317 \\
HuBERT \cite{hubert} & Masked prediction over clustered units & Raw audio & Self-supervised + supervised FT & Libri-Light; FT on~LibriSpeech & 60k + 960 & 316 \\
WavLM \cite{wavlm} & Masked prediction + denoising & Raw audio & Self-supervised & Libri-Light + GigaSpeech + VoxPopuli & 94k & 316 \\
MMS-1B \cite{mms} & wav2vec2.0-style SSL & Raw audio & Self-supervised & Multilingual speech corpus & $\sim$500k & 1000 \\
Qwen3-TTS-Tokenizer-12Hz \cite{Qwen3} & Neural audio codec reconstruction & Raw audio & Self-supervised & Multilingual speech corpus & $>$5M & 150 \\
Whisper Large-v3~\cite{whisper} & Sequence-to-sequence speech-to-text & Spectrogram & Supervised / pseudo-supervised & Web-scale speech--text pairs & $\sim$5M & 1550 \\
\midrule
AudioMAE \cite{audiomae} & Masked spectrogram reconstruction & Spectrogram & Self-supervised & AudioSet & $\sim$5.8k & 85.6 \\
WavJEPA-Nat \cite{wavjepa} & Latent representation prediction (JEPA) & Raw audio & Self-supervised & AudioSet (naturalistic scenes) & $\sim$4.8k & $\sim$200 \\
AST \cite{ast} & Audio event classification & Spectrogram & Supervised & AudioSet & $\sim$5.8k & 86.6 \\
CLAP \cite{clap} & Audio--text contrastive learning & Spectrogram & Cross-modal contrastive & LAION-Audio-630K + AudioSet & $\sim$10k & $\sim$400 \\
\midrule
emotion2vec+ Large~\cite{emotion2vec} & Multi-scale emotion learning & Raw audio & Pseudo-supervised & Large-scale emotional speech & $\sim$160k & $\sim$300 \\
OPERA-GT \cite{opera} & Generative reconstruction & Spectrogram & Self-supervised & Multi-source respiratory audio & 404 & 21 \\
\bottomrule
\end{tblr}
\end{table*}

\subsection{Speech Models}
\textbf{wav2vec~2.0}~\cite{wav2vec2} is a self-supervised speech representation model trained using a contrastive objective over discretized latent units. The large (LV-60) configuration is pretrained on approximately 60k hours of Libri-Light~\cite{librilight} and subsequently fine-tuned on 960 hours of LibriSpeech~\cite{librispeech} for automatic speech recognition. 

\textbf{HuBERT (Large)}~\cite{hubert} is a self-supervised speech model trained via masked prediction of offline cluster assignments derived from k-means clustering of acoustic features. Unlike wav2vec~2.0, it relies on iterative refinement of cluster targets rather than contrastive learning. The large variant is pretrained on 60k hours of Libri-Light~\cite{librilight} and fine-tuned on LibriSpeech~\cite{librispeech} for automatic speech recognition.

\textbf{WavLM (Large)}~\cite{wavlm} extends HuBERT by incorporating a denoising objective and training on mixtures of overlapping speech to improve robustness to real-world acoustic conditions. It combines masked prediction with augmentation strategies that simulate multi-speaker environments. The large configuration is pretrained on approximately 94k hours of speech spanning Libri-Light~\cite{librilight}, GigaSpeech~\cite{gigaspeech}, and VoxPopuli~\cite{voxpopuli}.

\textbf{MMS-1B}~\cite{mms} is a multilingual extension of wav2vec~2.0 designed to scale speech representation learning across languages. It is trained using self-supervised learning on approximately 500k hours of speech covering over 1,400 languages. We evaluate the 1B-parameter configuration.

\textbf{Qwen3-TTS-Tokenizer (12Hz)}~\cite{Qwen3} is a neural audio codec that encodes speech into discrete tokens at a fixed temporal resolution (12.5\,Hz) and reconstructs the waveform from these tokens. The model is trained using a self-supervised reconstruction objective on large-scale multilingual speech data as part of the Qwen3-TTS model.

\textbf{Whisper (Large-v3)}~\cite{whisper} is a sequence-to-sequence encoder–decoder model trained to map audio to text. It is trained on approximately 5 million hours of weakly and pseudo-labeled audio–text pairs, enabling robust performance across languages and domains. The model jointly learns multiple tasks including speech recognition, translation, and language identification through task conditioning. We evaluate the Large-v3 configuration.

\subsection{General Audio Models}
\textbf{AudioMAE}~\cite{audiomae} is a masked autoencoder that learns audio representations by reconstructing missing regions of log-mel spectrogram inputs. A large proportion of the input patches are masked during training, encouraging the model to capture global acoustic structure. The base configuration is pretrained on approximately 5.8k hours of audio from AudioSet~\cite{audioset}.

\textbf{WavJEPA-Nat}~\cite{wavjepa} is trained using a joint-embedding predictive architecture that learns to predict high-level latent representations of masked audio segments rather than reconstructing the input signal directly. The model is designed to capture semantic structure while being robust to low-level variations. It is pretrained on AudioSet~\cite{audioset} with additional naturalistic augmentations such as noise and reverberation. We use the naturalistic base variant.

\textbf{AST}~\cite{ast} adapts vision transformers to audio classification by operating on log-mel spectrogram patches. It leverages ImageNet-pretrained weights for initialization and is fine-tuned on AudioSet \cite{audioset} for supervised audio classification. This transfer from vision to audio enables strong performance with relatively limited audio data.

\textbf{CLAP}~\cite{clap} is a dual-encoder model trained to align audio and text representations using a contrastive objective. We evaluate a general-domain configuration trained on diverse audio--text pairs drawn from sources such as LAION-Audio~\cite{laion-audio} and AudioSet~\cite{audioset} with caption augmentation rather than variants specialized for music or speech.

\subsection{Domain-specific Models}
\textbf{emotion2vec+ (Large)}~\cite{emotion2vec} is designed to capture emotional characteristics in speech using a multi-stage training pipeline. It first learns representations from labeled emotional speech data and then scales to large unlabeled corpora via pseudo-labeling in a teacher–student framework. The large configuration is trained on emotional speech datasets totaling approximately 160k hours, with filtering applied during training.

\textbf{OPERA-GT}~\cite{opera} is a domain-specific model designed for respiratory sound analysis. It is trained using a self-supervised generative objective that reconstructs spectrogram representations, along with auxiliary tasks designed to capture clinically relevant acoustic features. The model is pretrained on approximately 400 hours of curated respiratory audio data. We use the generative (GT) variant.
\section{Implementation Details}
\label{app:implementation}
This section describes the implementation details of the \benchmarkname benchmark, including data augmentation, loss functions, class weighting, and hyperparameter optimization.

\subsection{Data Augmentation}
To improve robustness and partially compensate for limited dataset sizes, each training sample is augmented to produce additional training instances. Three augmentation strategies are applied jointly: (1) additive noise drawn from the Microsoft SNSD corpus~\cite{mssnsd} at a randomly sampled signal-to-noise ratio in $[0, 15]$\,dB; (2) convolutive reverberation using room impulse responses from the MIT IR Survey~\cite{mit-ir}; and (3) speed perturbation uniformly sampled from $[90\%, 110\%]$. For COVID-19 Sounds, speed perturbation is restricted to $[95\%, 105\%]$ to preserve the integrity of breathing patterns. Three augmented versions are produced per training sample for most datasets; this is increased to five for MDVR-KCL given its small sample count, and reduced to one for COVID-19 Sounds, where the full dataset is already used in training. Augmentation is applied exclusively to training data; validation and test sets are evaluated on clean audio only. All augmentation is implemented using SpeechBrain~\cite{speechbrain_v1, speechbrain}.

\subsection{Training Setup}
All binary and multi-label classification tasks are trained with binary cross-entropy loss, multi-class tasks with categorical cross-entropy loss, and regression tasks with weighted mean squared error (MSE). To mitigate the class imbalance that is prevalent across clinical datasets, all tasks use inverse-frequency weighting. For binary and multi-label tasks, positive class weights are computed per label as the ratio of negative to positive samples. For multi-class tasks, per-class weights are computed analogously. For regression tasks, sample weights are assigned via binning. Clinically established severity thresholds directly define the bins when available: MMSE scores are binned as severe ($\leq 9$), moderate ($10$--$18$), mild ($19$--$23$), and normal ($\geq 24$)~\cite{mmse}; and PHQ-8 scores are binned as $[0$--$4]$, $[5$--$9]$, $[10$--$14]$, $[15$--$19]$, and $[20+]$~\cite{phq8}. For tasks with discrete ordinal severity ratings (T11 and T14--T16), bin edges are derived from midpoints between consecutive unique values. Each sample is assigned a weight equal to the inverse frequency of its bin, normalized so that the mean per-sample weight equals 1 to maintain a stable loss scale across tasks.
 
\subsection{Hyperparameter Optimization}
For each model-task combination, hyperparameters are selected via Optuna~\cite{ozaki2025optunahub} over five trials to optimize validation loss. The learning rate is sampled log-uniformly over $[10^{-4}, 10^{-3}]$ and decayed linearly to one-tenth of its initial value over the training run. Weight decay (L2 regularization) is sampled log-uniformly over $[0.01, 0.1]$. All models are trained with a batch size of 16 in 32-bit floating-point precision for up to 50 epochs. Early stopping is triggered after 5 consecutive epochs without improvement in validation loss; the first 4 epochs are excluded from the early stopping criterion to allow sufficient warm-up. All experiments use a global random seed.

\section{Data Efficiency Analysis}
\label{app:data_efficiency}

\begin{figure}[!t]
    \centering
    \includegraphics[width=0.9\linewidth]{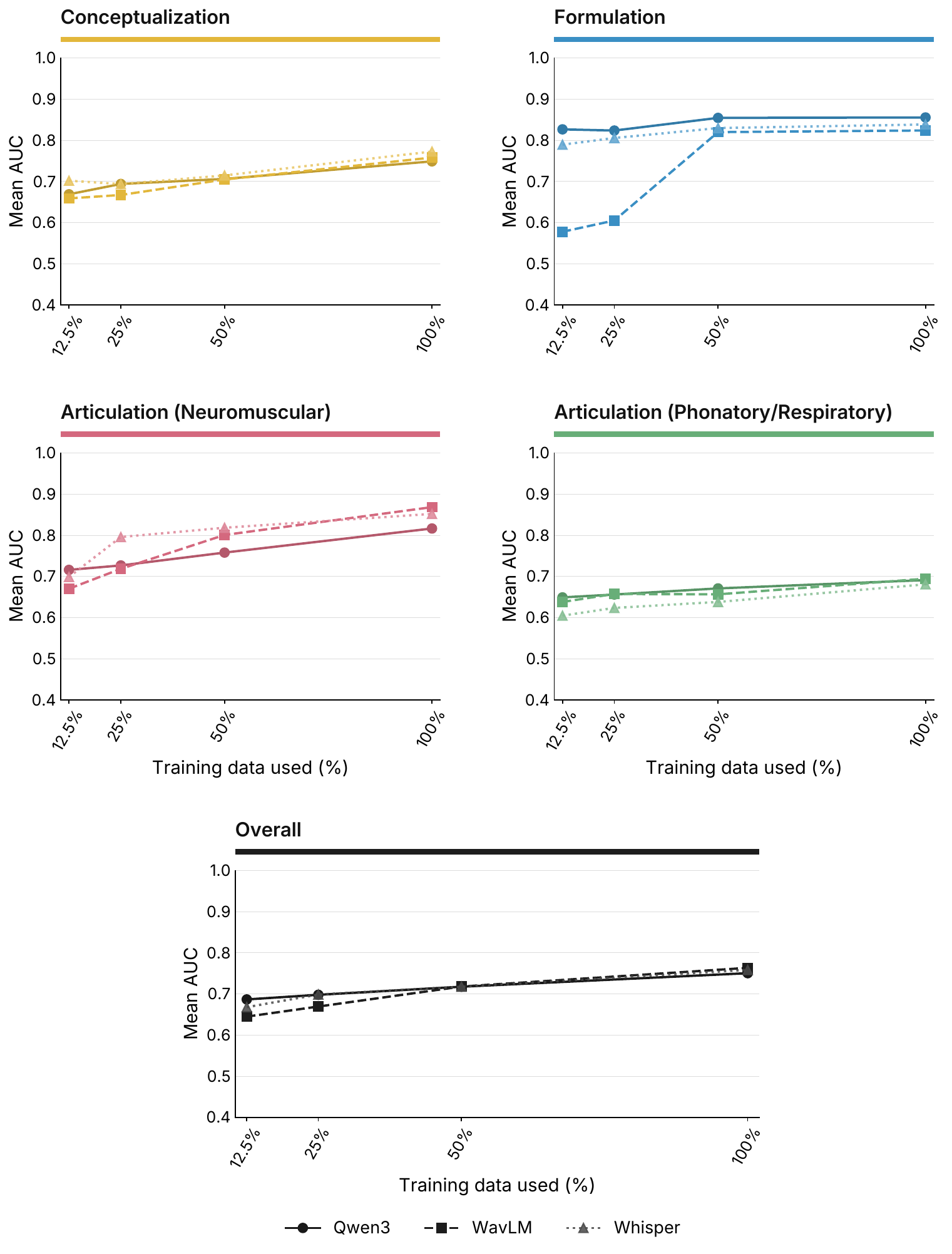}
    \caption{The data efficiency of Qwen3-TTS-Tokenizer \cite{Qwen3}, WavLM~\cite{wavlm}, and Whisper~\cite{whisper} with various amounts of training data.}
    \label{fig:data_efficiency}
\end{figure}

Clinical speech datasets are typically small, and label acquisition is costly. To characterize how different encoders perform under data scarcity, we train linear probes using 12.5\%, 25\%, 50\%, and 100\% of the available training data. This analysis is conducted for Qwen3-TTS-Tokenizer~\cite{Qwen3}, WavLM~\cite{wavlm}, and Whisper~\cite{whisper}, which were identified as the top-performing models in Section~\ref{sec:results-main}. For each data regime, training subsets are constructed using the same stratification protocol as the main experiments. This ensures that the label distribution is preserved across all data regimes. All models use identical subsampled splits at each fraction, and all probes are evaluated on the same test set. All per-task results are reported in \autoref{tab:data_efficiency}, while \autoref{fig:data_efficiency} summarizes AUC trajectories grouped by speech production stage.

Qwen3 emerges as the most data-efficient model, leading 11 out of 27 tasks when trained on the least amount of data. Whisper exhibits competitive performance at higher amounts of training data but demonstrates reduced effectiveness at the lowest fraction. WavLM exhibits substantial performance variability across tasks and data regimes; while it achieves strong performance on specific tasks at low amounts of training data (e.g., AUC: 0.54 for T1 at 12.5\%, AUC: 0.85 for T12 at 12.5\%), it also demonstrates severe performance degradation on others (e.g., AUC: 0.33 for T7 at 12.5\%). This variability precludes reliable deployment of WavLM in low-data regimes without task-specific validation.

At the task level, several conditions demonstrate strong performance with minimal training data. Aphasia detection (T9) achieves robust performance even at 12.5\% of training data, with Qwen3 reaching an AUC of 0.90 and improving to 0.97 at 100\%, indicating that aphasia-relevant speech patterns can be captured with limited supervision. Similarly, dysarthria detection (T10, T12), disfluency classification (T18), and multiple respiratory symptom tasks (T19, T20, T21, T23, T25, T26) exhibit relatively flat learning curves, with most models approaching within 0.05-0.10 AUC of their full-data performance using only 25\% of the training set. 

In contrast, Parkinson's detection tasks (T13, T14, T16) and Alzheimer's detection (T7) demonstrate the steepest data requirements. For T13, the AUC of Whisper improves from 0.61 to 0.73, while WavLM exhibits a more pronounced improvement from an AUC of 0.56 to 0.81. It is important to note that these proportions of training data reflect different absolute sample sizes across tasks: 12.5\% of training data corresponds to as few as 4 training samples for small datasets like MDVR-KCL (30 total subjects) and as many as 3,156 samples for larger datasets like COVID-19 Sounds (25,252 total samples), which partially accounts for the observed task-level variability in data efficiency.

These results expose substantial heterogeneity in data efficiency across both tasks and models. While the majority of tasks plateau by 25-50\% of available training data, specific conditions consistently require larger training sets across all evaluated encoders. Significant instability under data scarcity indicates that current self-supervised speech encoders lack robustness when training supervision is limited. Overall, data efficiency depends critically on the interaction between encoder architecture, pretraining strategy, and the specific acoustic manifestations of each clinical condition.

\begin{table*}[t]
\centering
\caption{Data-efficiency results. Per-task performance for Qwen3, WavLM, and Whisper when trained on 12.5\%, 25\%, 50\%, and 100\% of the available training data. Each cell reports the metric value with the bootstrapped 95\% confidence interval below. The best of the three models for each task and training fraction is shown in bold.}
\label{tab:data_efficiency}
\resizebox{\textwidth}{!}{%
\begin{tabular}{llcccccccccccc}
\toprule
\textbf{Task} & \textbf{Metric} & \multicolumn{3}{c}{\textbf{12.5\%}} & \multicolumn{3}{c}{\textbf{25\%}} & \multicolumn{3}{c}{\textbf{50\%}} & \multicolumn{3}{c}{\textbf{100\%}} \\
\cmidrule(lr){3-5} \cmidrule(lr){6-8} \cmidrule(lr){9-11} \cmidrule(lr){12-14}
 & & Qwen3 & WavLM & Whisper & Qwen3 & WavLM & Whisper & Qwen3 & WavLM & Whisper & Qwen3 & WavLM & Whisper \\
\midrule
\multicolumn{14}{l}{\textbf{Conceptualization}} \\[4pt]
T1 (Depression detection) & AUC & 0.43 & \textbf{0.54} & 0.47 & 0.46 & \textbf{0.49} & 0.38 & 0.44 & \textbf{0.47} & 0.39 & 0.46 & \textbf{0.57} & 0.55 \\
 & & (0.27, 0.59) & (0.36, 0.70) & (0.29, 0.63) & (0.29, 0.63) & (0.32, 0.65) & (0.20, 0.55) & (0.26, 0.61) & (0.30, 0.63) & (0.21, 0.56) & (0.29, 0.63) & (0.40, 0.73) & (0.37, 0.72) \\
T2 (Depression severity) & MAE & 5.34 & 7.09 & \textbf{5.26} & \textbf{5.25} & 6.62 & 5.25 & \textbf{5.42} & 5.84 & 5.48 & \textbf{5.31} & 5.37 & 5.34 \\
 & & (4.34, 6.43) & (5.67, 8.73) & (4.31, 6.23) & (4.32, 6.23) & (5.28, 8.17) & (4.32, 6.16) & (4.34, 6.59) & (4.74, 7.10) & (4.45, 6.59) & (4.30, 6.38) & (4.40, 6.37) & (4.35, 6.35) \\
T3 (Emotion classification) & AUC & 0.74 & 0.72 & \textbf{0.81} & 0.78 & 0.76 & \textbf{0.82} & 0.84 & 0.81 & \textbf{0.85} & \textbf{0.89} & 0.86 & 0.87 \\
 & & (0.70, 0.77) & (0.70, 0.74) & (0.80, 0.83) & (0.74, 0.83) & (0.72, 0.80) & (0.80, 0.84) & (0.82, 0.86) & (0.79, 0.84) & (0.83, 0.88) & (0.86, 0.91) & (0.84, 0.88) & (0.85, 0.90) \\
T4 (Binary emotion classification) & AUC & 0.62 & 0.59 & \textbf{0.72} & 0.68 & 0.65 & \textbf{0.76} & 0.66 & 0.70 & \textbf{0.76} & 0.78 & 0.76 & \textbf{0.82} \\
 & & (0.58, 0.67) & (0.48, 0.70) & (0.68, 0.75) & (0.60, 0.75) & (0.54, 0.76) & (0.72, 0.79) & (0.64, 0.68) & (0.61, 0.80) & (0.73, 0.79) & (0.75, 0.81) & (0.75, 0.78) & (0.79, 0.84) \\
T5 (Emotion classification) & AUC & 0.81 & 0.78 & \textbf{0.82} & 0.81 & 0.78 & \textbf{0.82} & 0.82 & 0.81 & \textbf{0.84} & 0.83 & 0.83 & \textbf{0.86} \\
 & & (0.78, 0.84) & (0.76, 0.81) & (0.80, 0.84) & (0.78, 0.84) & (0.76, 0.81) & (0.80, 0.84) & (0.80, 0.85) & (0.78, 0.84) & (0.81, 0.87) & (0.81, 0.86) & (0.81, 0.85) & (0.85, 0.88) \\
T6 (Binary emotion classification) & AUC & \textbf{0.75} & 0.66 & 0.69 & \textbf{0.75} & 0.66 & 0.69 & \textbf{0.77} & 0.73 & 0.73 & \textbf{0.78} & 0.76 & 0.76 \\
 & & (0.70, 0.79) & (0.62, 0.70) & (0.68, 0.71) & (0.70, 0.79) & (0.62, 0.70) & (0.68, 0.71) & (0.74, 0.80) & (0.70, 0.76) & (0.70, 0.76) & (0.75, 0.81) & (0.73, 0.80) & (0.73, 0.79) \\
\midrule
\multicolumn{14}{l}{\textbf{Formulation}} \\[4pt]
T7 (Alzheimer's detection) & AUC & 0.75 & 0.33 & \textbf{0.76} & \textbf{0.75} & 0.31 & 0.73 & \textbf{0.76} & 0.69 & 0.75 & 0.74 & 0.69 & \textbf{0.75} \\
 & & (0.59, 0.89) & (0.17, 0.50) & (0.61, 0.90) & (0.59, 0.89) & (0.16, 0.49) & (0.56, 0.87) & (0.59, 0.90) & (0.53, 0.84) & (0.59, 0.89) & (0.58, 0.88) & (0.53, 0.83) & (0.59, 0.89) \\
T8 (Alzheimer's severity) & MAE & \textbf{13.44} & 21.92 & 18.87 & \textbf{12.02} & 18.91 & 13.89 & 11.08 & 13.61 & \textbf{9.67} & 10.75 & \textbf{8.58} & 9.97 \\
 & & (11.68, 15.04) & (20.31, 23.37) & (17.21, 20.37) & (10.35, 13.56) & (17.25, 20.44) & (12.15, 15.42) & (9.44, 12.68) & (11.93, 15.16) & (8.29, 10.98) & (9.25, 12.20) & (7.23, 9.87) & (8.60, 11.33) \\
T9 (Aphasia detection) & AUC & \textbf{0.90} & 0.83 & 0.82 & 0.90 & \textbf{0.90} & 0.89 & \textbf{0.95} & 0.95 & 0.91 & \textbf{0.97} & 0.96 & 0.92 \\
 & & (0.81, 0.98) & (0.69, 0.94) & (0.68, 0.93) & (0.79, 0.99) & (0.79, 0.99) & (0.78, 0.97) & (0.89, 0.99) & (0.89, 0.99) & (0.82, 0.98) & (0.92, 1.00) & (0.91, 0.99) & (0.84, 0.98) \\
\midrule
\multicolumn{14}{l}{\textbf{Articulation (Neuromuscular)}} \\[4pt]
T10 (Dysarthria detection) & AUC & \textbf{0.79} & 0.59 & 0.61 & 0.68 & 0.66 & \textbf{0.84} & 0.75 & 0.75 & \textbf{0.87} & 0.87 & 0.88 & \textbf{0.91} \\
 & & (0.67, 0.91) & (0.44, 0.75) & (0.38, 0.84) & (0.37, 0.99) & (0.48, 0.85) & (0.71, 0.97) & (0.44, 1.06) & (0.48, 1.03) & (0.75, 1.00) & (0.77, 0.97) & (0.76, 1.00) & (0.84, 0.98) \\
T11 (Dysarthria severity) & MAE & \textbf{0.56} & 0.57 & 0.61 & 0.49 & 0.54 & \textbf{0.37} & 0.61 & 0.61 & \textbf{0.47} & 0.54 & 0.63 & \textbf{0.43} \\
 & & (0.36, 0.76) & (0.35, 0.78) & (0.32, 0.89) & (0.33, 0.65) & (0.33, 0.74) & (0.27, 0.48) & (0.31, 0.92) & (0.38, 0.84) & (0.29, 0.66) & (0.33, 0.75) & (0.37, 0.90) & (0.27, 0.59) \\
T12 (Dysarthria detection) & AUC & \textbf{0.93} & 0.85 & 0.90 & 0.92 & 0.90 & \textbf{0.94} & 0.92 & 0.91 & \textbf{0.95} & 0.95 & 0.96 & \textbf{0.97} \\
 & & (0.87, 0.99) & (0.63, 1.06) & (0.84, 0.96) & (0.85, 0.98) & (0.80, 1.01) & (0.87, 1.00) & (0.81, 1.03) & (0.76, 1.05) & (0.89, 1.01) & (0.90, 0.99) & (0.91, 1.01) & (0.92, 1.01) \\
T13 (Parkinson's detection) & AUC & 0.55 & 0.56 & \textbf{0.61} & 0.60 & 0.56 & \textbf{0.73} & 0.61 & \textbf{0.74} & 0.72 & 0.67 & \textbf{0.81} & 0.73 \\
 & & (0.37, 0.74) & (0.40, 0.72) & (0.50, 0.71) & (0.41, 0.80) & (0.45, 0.68) & (0.66, 0.81) & (0.52, 0.70) & (0.61, 0.86) & (0.63, 0.80) & (0.51, 0.83) & (0.75, 0.88) & (0.65, 0.81) \\
T14 (Parkinson's severity) & MAE & 0.45 & \textbf{0.44} & 0.46 & \textbf{0.43} & 0.44 & 0.46 & \textbf{0.44} & 0.45 & 0.45 & \textbf{0.42} & 0.44 & 0.43 \\
 & & (0.39, 0.50) & (0.39, 0.49) & (0.43, 0.49) & (0.37, 0.49) & (0.40, 0.48) & (0.43, 0.48) & (0.40, 0.48) & (0.42, 0.47) & (0.42, 0.48) & (0.36, 0.49) & (0.42, 0.47) & (0.39, 0.46) \\
T15 (Parkinson's severity) & MAE & 0.45 & \textbf{0.44} & 0.46 & 0.45 & \textbf{0.44} & 0.46 & \textbf{0.44} & 0.44 & 0.46 & \textbf{0.42} & 0.44 & 0.44 \\
 & & (0.39, 0.51) & (0.38, 0.50) & (0.42, 0.50) & (0.39, 0.52) & (0.37, 0.50) & (0.43, 0.50) & (0.35, 0.52) & (0.38, 0.50) & (0.42, 0.50) & (0.33, 0.51) & (0.38, 0.49) & (0.39, 0.49) \\
T16 (Parkinson's severity) & MAE & 0.45 & \textbf{0.45} & 0.47 & 0.46 & \textbf{0.45} & 0.47 & 0.46 & \textbf{0.46} & 0.46 & \textbf{0.41} & 0.45 & 0.44 \\
 & & (0.41, 0.50) & (0.40, 0.50) & (0.45, 0.50) & (0.41, 0.51) & (0.40, 0.50) & (0.44, 0.50) & (0.40, 0.52) & (0.42, 0.49) & (0.43, 0.49) & (0.34, 0.48) & (0.41, 0.48) & (0.41, 0.46) \\
T17 (Disfluency detection) & AUC & 0.68 & 0.71 & \textbf{0.74} & 0.76 & 0.77 & \textbf{0.77} & 0.77 & \textbf{0.82} & 0.79 & 0.81 & \textbf{0.86} & 0.85 \\
 & & (0.56, 0.80) & (0.65, 0.78) & (0.69, 0.78) & (0.66, 0.87) & (0.68, 0.85) & (0.71, 0.83) & (0.72, 0.83) & (0.79, 0.86) & (0.75, 0.83) & (0.75, 0.86) & (0.83, 0.88) & (0.81, 0.89) \\
T18 (Disfluency classification) & AUC & 0.62 & \textbf{0.64} & 0.64 & 0.67 & 0.70 & \textbf{0.70} & 0.74 & \textbf{0.78} & 0.76 & 0.78 & \textbf{0.84} & 0.81 \\
 & & (0.60, 0.65) & (0.61, 0.67) & (0.57, 0.70) & (0.63, 0.71) & (0.66, 0.74) & (0.68, 0.72) & (0.71, 0.76) & (0.77, 0.79) & (0.75, 0.78) & (0.78, 0.79) & (0.82, 0.85) & (0.78, 0.83) \\
\midrule
\multicolumn{14}{l}{\textbf{Articulation (Phonatory / Respiratory)}} \\[4pt]
T19 (Respiratory symptom detection) & AUC & \textbf{0.58} & 0.57 & 0.55 & 0.62 & \textbf{0.63} & 0.57 & \textbf{0.64} & 0.62 & 0.58 & \textbf{0.69} & 0.68 & 0.68 \\
 & & (0.55, 0.61) & (0.54, 0.61) & (0.51, 0.58) & (0.58, 0.65) & (0.60, 0.66) & (0.54, 0.61) & (0.60, 0.68) & (0.59, 0.66) & (0.55, 0.62) & (0.66, 0.72) & (0.65, 0.71) & (0.65, 0.71) \\
T20 (Respiratory symptom detection) & AUC & \textbf{0.59} & 0.58 & 0.56 & 0.59 & 0.59 & \textbf{0.60} & 0.61 & \textbf{0.62} & 0.62 & 0.64 & 0.65 & \textbf{0.65} \\
 & & (0.58, 0.60) & (0.56, 0.59) & (0.55, 0.58) & (0.58, 0.60) & (0.58, 0.61) & (0.58, 0.61) & (0.60, 0.62) & (0.60, 0.63) & (0.60, 0.63) & (0.62, 0.65) & (0.64, 0.66) & (0.64, 0.66) \\
T21 (COVID-19 detection) & AUC & \textbf{0.64} & 0.54 & 0.49 & \textbf{0.60} & 0.56 & 0.50 & \textbf{0.59} & 0.49 & 0.49 & 0.61 & \textbf{0.65} & 0.61 \\
 & & (0.56, 0.72) & (0.46, 0.62) & (0.41, 0.57) & (0.52, 0.69) & (0.48, 0.64) & (0.41, 0.58) & (0.50, 0.67) & (0.41, 0.56) & (0.40, 0.57) & (0.52, 0.69) & (0.56, 0.72) & (0.53, 0.69) \\
T22 (COVID-19 detection) & AUC & 0.63 & \textbf{0.64} & 0.60 & 0.64 & \textbf{0.66} & 0.65 & 0.61 & \textbf{0.65} & 0.62 & 0.65 & 0.68 & \textbf{0.70} \\
 & & (0.57, 0.67) & (0.58, 0.69) & (0.55, 0.65) & (0.59, 0.69) & (0.61, 0.71) & (0.59, 0.69) & (0.56, 0.66) & (0.60, 0.69) & (0.57, 0.66) & (0.59, 0.70) & (0.63, 0.73) & (0.65, 0.74) \\
T23 (Respiratory symptom classification) & AUC & \textbf{0.57} & 0.57 & 0.56 & \textbf{0.59} & 0.58 & 0.58 & 0.60 & 0.60 & \textbf{0.60} & 0.60 & 0.60 & \textbf{0.61} \\
 & & (0.57, 0.58) & (0.56, 0.58) & (0.55, 0.57) & (0.58, 0.59) & (0.57, 0.59) & (0.57, 0.58) & (0.59, 0.60) & (0.59, 0.60) & (0.60, 0.61) & (0.59, 0.61) & (0.59, 0.61) & (0.61, 0.62) \\
T24 (Respiratory symptom detection) & AUC & 0.68 & 0.68 & \textbf{0.69} & 0.69 & 0.70 & \textbf{0.71} & 0.71 & 0.71 & \textbf{0.72} & \textbf{0.73} & 0.72 & 0.72 \\
 & & (0.63, 0.72) & (0.64, 0.72) & (0.65, 0.74) & (0.64, 0.73) & (0.66, 0.74) & (0.67, 0.75) & (0.67, 0.76) & (0.67, 0.75) & (0.67, 0.76) & (0.69, 0.77) & (0.68, 0.76) & (0.68, 0.76) \\
T25 (COVID-19 detection) & AUC & \textbf{0.76} & 0.75 & 0.73 & 0.73 & 0.74 & \textbf{0.75} & \textbf{0.78} & 0.74 & 0.76 & \textbf{0.79} & 0.77 & 0.77 \\
 & & (0.71, 0.81) & (0.70, 0.79) & (0.68, 0.77) & (0.68, 0.78) & (0.68, 0.78) & (0.70, 0.79) & (0.73, 0.83) & (0.69, 0.78) & (0.71, 0.80) & (0.74, 0.83) & (0.72, 0.81) & (0.72, 0.81) \\
T26 (Respiratory symptom classification) & AUC & 0.55 & \textbf{0.55} & 0.54 & 0.55 & \textbf{0.56} & 0.55 & \textbf{0.58} & 0.57 & 0.57 & \textbf{0.60} & 0.59 & 0.59 \\
 & & (0.51, 0.58) & (0.51, 0.58) & (0.51, 0.57) & (0.52, 0.59) & (0.53, 0.60) & (0.52, 0.58) & (0.54, 0.61) & (0.53, 0.61) & (0.53, 0.60) & (0.57, 0.63) & (0.56, 0.63) & (0.55, 0.62) \\
T27 (Vocal pathology detection) & AUC & 0.85 & \textbf{0.87} & 0.72 & 0.89 & \textbf{0.90} & 0.72 & 0.92 & \textbf{0.92} & 0.79 & \textbf{0.92} & 0.91 & 0.81 \\
 & & (0.80, 0.90) & (0.83, 0.92) & (0.67, 0.77) & (0.85, 0.93) & (0.86, 0.94) & (0.65, 0.78) & (0.87, 0.95) & (0.87, 0.96) & (0.73, 0.85) & (0.87, 0.95) & (0.87, 0.95) & (0.75, 0.86) \\
\bottomrule
\end{tabular}
}
\end{table*}
\section{Zero-shot Transfer Analysis}
\label{app:zero-shot}
Beyond the aggregate cross-category patterns reported in Section~\ref{sec:zero-shot}, individual task-level transfer pairs reveal several insights into which clinical manifestations produce generalizable representations and which tasks exhibit fragmentation. Table~\ref{tab:zero-shot-all} reports complete zero-shot transfer results for all model-source-target combinations.

\paragraph{Transfer Asymmetries Within Production Stages.} Bidirectional transfer within the same production stage reveals which task representations are more generalizable. Alzheimer's $\rightarrow$ Aphasia (T7 $\rightarrow$ T9) achieves an AUC of 0.94 (HuBERT) while Aphasia $\rightarrow$ Alzheimer's (T9 $\rightarrow$ T7) reaches only 0.74 (Whisper), indicating that dementia-trained representations capture broader language disruption patterns that extend to aphasia, whereas aphasia-specific features are less informative for Alzheimer's detection. Similarly, dysarthria transfer is asymmetric: TORGO $\rightarrow$ UASpeech (T10 $\rightarrow$ T12) achieves an AUC of 0.92 (Whisper) while UASpeech $\rightarrow$ TORGO (T12 $\rightarrow$ T10) reaches only 0.76. TORGO includes speakers with more severe dysarthria (primarily cerebral palsy), while UASpeech covers a broader range of severities.
These asymmetries suggest that certain clinical manifestations yield more generalizable acoustic signatures than others, though the mechanisms differ: severity in the case of dysarthria, and potentially the involvement of multiple linguistic domains in the case of dementia, versus focal language impairment in aphasia.



\paragraph{COVID-19 and Respiratory Task Fragmentation.} COVID-19 and respiratory tasks show minimal cross-transfer. COVID-19 Sounds $\rightarrow$ Coswara achieves a maximum AUC of 0.79, within-dataset tasks (T21 $\rightarrow$ T22 and T22 $\rightarrow$ T21) achieve a maximum AUC of 0.64, and general respiratory symptoms (T19, T24) transfer to COVID-19 detection (T21, T22, T25) with AUC $\leq$ 0.69. This fragmentation reflects differences in recording protocols (cough vs. sustained phonation) or population characteristics, and that current encoders fail to extract protocol-invariant respiratory features.

\paragraph{Sample Size versus Feature Alignment in Emotion $\rightarrow$ Depression Transfer.} Emotion-labeled data provides modest benefits for depression detection, likely through sample size rather than feature alignment. Emotion $\rightarrow$ Depression transfer (maximum AUC: 0.75) exceeds performance when trained and tested for depression itself (maximum AUC: 0.65), but the gain is small relative to the multi-fold increase in training samples (4,246 vs. 163), indicating that general affective prosody and clinical depression capture partially overlapping but distinct acoustic patterns.

\paragraph{Cross-category Transfer Reveals Acoustic Hierarchy.} Phonatory / Respiratory $\rightarrow$ Conceptualization (maximum AUC: 0.83) and Phonatory / Respiratory $\rightarrow$ Formulation (maximum AUC: 0.88) succeed, while reverse transfers fail (AUC $\leq$ 0.60). Conceptualization tasks produce the weakest cross-category transfer as sources (AUC $\leq$ 0.67). This asymmetry suggests that low-level acoustic features, such as voice quality and phonation characteristics, provide useful priors for higher-level cognitive-linguistic tasks, raising the possibility that multi-stage screening pipelines could leverage simpler phonatory tasks as initial filters before deploying more specialized models for cognitive assessment.

\begin{sidewaystable}[htbp]
\centering
\scriptsize
\caption{Zero-shot transfer performance within and across categories, reported as AUC with 95\% confidence intervals.}
\label{tab:zero-shot-all}
\resizebox{\textwidth}{!}{
\begin{tabular}{c >{\raggedright\arraybackslash}p{0.1\linewidth} c >{\raggedright\arraybackslash}p{0.1\linewidth} >{\centering\arraybackslash}p{0.074\linewidth} >{\centering\arraybackslash}p{0.074\linewidth} >{\centering\arraybackslash}p{0.074\linewidth} >{\centering\arraybackslash}p{0.074\linewidth} >{\centering\arraybackslash}p{0.074\linewidth} >{\centering\arraybackslash}p{0.074\linewidth} >{\centering\arraybackslash}p{0.074\linewidth} >{\centering\arraybackslash}p{0.074\linewidth} >{\centering\arraybackslash}p{0.074\linewidth} >{\centering\arraybackslash}p{0.074\linewidth} >{\centering\arraybackslash}p{0.074\linewidth} >{\centering\arraybackslash}p{0.074\linewidth}}
\toprule
\textbf{Source Task ID} & \textbf{Source Task} & \textbf{Target Task ID} & \textbf{Target Task} & \textbf{Qwen3} & \textbf{WavLM} & \textbf{AST} & \textbf{AudioMAE} & \textbf{CLAP} & \textbf{emotion2vec+} & \textbf{HuBERT} & \textbf{MMS} & \textbf{OPERA-GT} & \textbf{wav2vec 2.0} & \textbf{WavJEPA} & \textbf{Whisper} \\
\midrule
\multicolumn{16}{l}{\textbf{Conceptualization}} \\
\midrule
T4 & Emotion Classification & T6 & Emotion Classification & 0.49 \scriptsize{(0.48, 0.50)} & 0.49 \scriptsize{(0.46, 0.52)} & 0.57 \scriptsize{(0.54, 0.59)} & 0.54 \scriptsize{(0.52, 0.56)} & 0.54 \scriptsize{(0.52, 0.56)} & \textbf{0.86} \scriptsize{(0.84, 0.87)} & 0.44 \scriptsize{(0.41, 0.47)} & 0.49 \scriptsize{(0.47, 0.51)} & 0.56 \scriptsize{(0.54, 0.57)} & 0.37 \scriptsize{(0.34, 0.40)} & 0.55 \scriptsize{(0.54, 0.57)} & 0.58 \scriptsize{(0.56, 0.60)} \\
T6 & Emotion Classification & T4 & Emotion Classification & 0.56 \scriptsize{(0.52, 0.60)} & 0.54 \scriptsize{(0.48, 0.60)} & 0.58 \scriptsize{(0.55, 0.62)} & 0.65 \scriptsize{(0.61, 0.69)} & 0.56 \scriptsize{(0.53, 0.59)} & \textbf{0.82} \scriptsize{(0.79, 0.84)} & 0.59 \scriptsize{(0.54, 0.63)} & 0.44 \scriptsize{(0.40, 0.48)} & 0.51 \scriptsize{(0.47, 0.55)} & 0.41 \scriptsize{(0.38, 0.43)} & 0.62 \scriptsize{(0.59, 0.65)} & 0.71 \scriptsize{(0.67, 0.75)} \\
T6 & Emotion Classification & T1 & Depression Detection & 0.58 \scriptsize{(0.50, 0.65)} & 0.50 \scriptsize{(0.43, 0.58)} & 0.41 \scriptsize{(0.34, 0.48)} & 0.44 \scriptsize{(0.36, 0.52)} & 0.43 \scriptsize{(0.36, 0.52)} & \textbf{0.75} \scriptsize{(0.68, 0.81)} & 0.56 \scriptsize{(0.48, 0.64)} & 0.43 \scriptsize{(0.35, 0.51)} & 0.45 \scriptsize{(0.37, 0.53)} & 0.54 \scriptsize{(0.46, 0.63)} & 0.58 \scriptsize{(0.50, 0.65)} & 0.50 \scriptsize{(0.43, 0.57)} \\
T4 & Emotion Classification & T1 & Depression Detection & 0.38 \scriptsize{(0.31, 0.46)} & 0.41 \scriptsize{(0.33, 0.50)} & 0.39 \scriptsize{(0.31, 0.46)} & 0.39 \scriptsize{(0.32, 0.47)} & 0.42 \scriptsize{(0.34, 0.49)} & \textbf{0.74} \scriptsize{(0.67, 0.80)} & 0.47 \scriptsize{(0.39, 0.56)} & 0.58 \scriptsize{(0.49, 0.66)} & 0.40 \scriptsize{(0.32, 0.48)} & 0.46 \scriptsize{(0.38, 0.55)} & 0.57 \scriptsize{(0.49, 0.65)} & 0.40 \scriptsize{(0.33, 0.47)} \\
T1 & Depression Detection & T4 & Emotion Classification & 0.35 \scriptsize{(0.31, 0.38)} & 0.56 \scriptsize{(0.51, 0.60)} & 0.43 \scriptsize{(0.40, 0.46)} & 0.45 \scriptsize{(0.42, 0.48)} & 0.53 \scriptsize{(0.49, 0.56)} & \textbf{0.93} \scriptsize{(0.90, 0.95)} & 0.40 \scriptsize{(0.36, 0.44)} & 0.44 \scriptsize{(0.40, 0.48)} & 0.42 \scriptsize{(0.38, 0.45)} & 0.54 \scriptsize{(0.50, 0.58)} & 0.49 \scriptsize{(0.47, 0.52)} & 0.49 \scriptsize{(0.44, 0.54)} \\
T1 & Depression Detection & T6 & Emotion Classification & 0.58 \scriptsize{(0.56, 0.60)} & 0.54 \scriptsize{(0.51, 0.56)} & 0.45 \scriptsize{(0.42, 0.47)} & 0.54 \scriptsize{(0.51, 0.56)} & 0.52 \scriptsize{(0.50, 0.54)} & \textbf{0.82} \scriptsize{(0.79, 0.84)} & 0.61 \scriptsize{(0.58, 0.64)} & 0.45 \scriptsize{(0.43, 0.47)} & 0.43 \scriptsize{(0.41, 0.45)} & 0.55 \scriptsize{(0.53, 0.57)} & 0.53 \scriptsize{(0.50, 0.55)} & 0.59 \scriptsize{(0.56, 0.62)} \\
\midrule
\multicolumn{16}{l}{\textbf{Formulation}} \\
\midrule
T9 & Aphasia Detection & T7 & Alzheimer's Detection & 0.70 \scriptsize{(0.64, 0.76)} & 0.74 \scriptsize{(0.68, 0.80)} & 0.65 \scriptsize{(0.58, 0.71)} & 0.65 \scriptsize{(0.58, 0.72)} & 0.57 \scriptsize{(0.50, 0.63)} & 0.48 \scriptsize{(0.41, 0.55)} & 0.73 \scriptsize{(0.67, 0.79)} & 0.74 \scriptsize{(0.68, 0.80)} & 0.66 \scriptsize{(0.60, 0.72)} & 0.69 \scriptsize{(0.62, 0.74)} & 0.68 \scriptsize{(0.62, 0.73)} & \textbf{0.74} \scriptsize{(0.68, 0.80)} \\
T7 & Alzheimer's Detection & T9 & Aphasia Detection & 0.82 \scriptsize{(0.77, 0.87)} & 0.90 \scriptsize{(0.86, 0.93)} & 0.84 \scriptsize{(0.79, 0.89)} & 0.84 \scriptsize{(0.78, 0.88)} & 0.67 \scriptsize{(0.59, 0.75)} & 0.61 \scriptsize{(0.56, 0.66)} & \textbf{0.94} \scriptsize{(0.92, 0.96)} & 0.83 \scriptsize{(0.78, 0.88)} & 0.88 \scriptsize{(0.83, 0.91)} & 0.78 \scriptsize{(0.72, 0.84)} & 0.83 \scriptsize{(0.77, 0.88)} & 0.88 \scriptsize{(0.83, 0.92)} \\
\midrule
\multicolumn{16}{l}{\textbf{Articulation (Neuromuscular)}} \\
\midrule
T10 & Dysarthria Detection & T17 & Disfluency Detection & 0.60 \scriptsize{(0.48, 0.68)} & \textbf{0.73} \scriptsize{(0.65, 0.80)} & 0.59 \scriptsize{(0.50, 0.67)} & 0.57 \scriptsize{(0.45, 0.66)} & 0.51 \scriptsize{(0.41, 0.64)} & 0.56 \scriptsize{(0.53, 0.60)} & 0.71 \scriptsize{(0.67, 0.76)} & 0.70 \scriptsize{(0.62, 0.77)} & 0.63 \scriptsize{(0.55, 0.70)} & 0.66 \scriptsize{(0.61, 0.71)} & 0.70 \scriptsize{(0.60, 0.78)} & 0.73 \scriptsize{(0.66, 0.77)} \\
T12 & Dysarthria Detection & T17 & Disfluency Detection & 0.66 \scriptsize{(0.58, 0.72)} & 0.64 \scriptsize{(0.57, 0.70)} & 0.62 \scriptsize{(0.54, 0.69)} & 0.60 \scriptsize{(0.52, 0.67)} & 0.50 \scriptsize{(0.40, 0.62)} & 0.60 \scriptsize{(0.55, 0.64)} & 0.57 \scriptsize{(0.52, 0.61)} & 0.57 \scriptsize{(0.44, 0.66)} & 0.61 \scriptsize{(0.51, 0.69)} & 0.65 \scriptsize{(0.60, 0.69)} & 0.68 \scriptsize{(0.60, 0.74)} & \textbf{0.70} \scriptsize{(0.65, 0.74)} \\
T10 & Dysarthria Detection & T12 & Dysarthria Detection & 0.83 \scriptsize{(0.72, 0.92)} & 0.90 \scriptsize{(0.83, 0.96)} & 0.78 \scriptsize{(0.63, 0.90)} & 0.78 \scriptsize{(0.62, 0.90)} & 0.27 \scriptsize{(0.16, 0.39)} & 0.76 \scriptsize{(0.65, 0.86)} & 0.91 \scriptsize{(0.83, 0.96)} & 0.80 \scriptsize{(0.68, 0.90)} & 0.55 \scriptsize{(0.41, 0.68)} & 0.82 \scriptsize{(0.72, 0.91)} & 0.52 \scriptsize{(0.39, 0.64)} & \textbf{0.92} \scriptsize{(0.84, 0.98)} \\
T12 & Dysarthria Detection & T13 & Parkinson's Detection & 0.56 \scriptsize{(0.41, 0.70)} & 0.67 \scriptsize{(0.56, 0.79)} & 0.54 \scriptsize{(0.37, 0.73)} & \textbf{0.83} \scriptsize{(0.70, 0.93)} & 0.73 \scriptsize{(0.59, 0.85)} & 0.67 \scriptsize{(0.54, 0.80)} & 0.67 \scriptsize{(0.51, 0.83)} & 0.61 \scriptsize{(0.50, 0.72)} & 0.30 \scriptsize{(0.16, 0.47)} & 0.55 \scriptsize{(0.48, 0.62)} & 0.79 \scriptsize{(0.66, 0.88)} & 0.65 \scriptsize{(0.56, 0.73)} \\
T12 & Dysarthria Detection & T10 & Dysarthria Detection & 0.62 \scriptsize{(0.45, 0.81)} & 0.72 \scriptsize{(0.54, 0.89)} & 0.66 \scriptsize{(0.48, 0.84)} & 0.63 \scriptsize{(0.45, 0.81)} & 0.62 \scriptsize{(0.46, 0.78)} & 0.71 \scriptsize{(0.60, 0.82)} & 0.68 \scriptsize{(0.53, 0.84)} & 0.60 \scriptsize{(0.41, 0.80)} & 0.51 \scriptsize{(0.36, 0.67)} & 0.71 \scriptsize{(0.59, 0.86)} & 0.71 \scriptsize{(0.58, 0.86)} & \textbf{0.76} \scriptsize{(0.58, 0.93)} \\
T10 & Dysarthria Detection & T13 & Parkinson's Detection & 0.50 \scriptsize{(0.34, 0.65)} & 0.52 \scriptsize{(0.36, 0.68)} & 0.46 \scriptsize{(0.28, 0.65)} & 0.65 \scriptsize{(0.48, 0.80)} & 0.71 \scriptsize{(0.54, 0.85)} & 0.42 \scriptsize{(0.25, 0.60)} & 0.56 \scriptsize{(0.41, 0.72)} & 0.56 \scriptsize{(0.40, 0.73)} & 0.45 \scriptsize{(0.28, 0.63)} & 0.54 \scriptsize{(0.43, 0.65)} & 0.51 \scriptsize{(0.34, 0.67)} & \textbf{0.72} \scriptsize{(0.57, 0.85)} \\
T13 & Parkinson's Detection & T12 & Dysarthria Detection & 0.47 \scriptsize{(0.32, 0.62)} & 0.62 \scriptsize{(0.53, 0.71)} & 0.71 \scriptsize{(0.58, 0.81)} & 0.53 \scriptsize{(0.39, 0.67)} & \textbf{0.81} \scriptsize{(0.75, 0.87)} & 0.70 \scriptsize{(0.62, 0.77)} & 0.74 \scriptsize{(0.69, 0.79)} & 0.26 \scriptsize{(0.18, 0.32)} & 0.28 \scriptsize{(0.17, 0.40)} & 0.35 \scriptsize{(0.26, 0.44)} & 0.58 \scriptsize{(0.46, 0.70)} & 0.48 \scriptsize{(0.36, 0.58)} \\
T13 & Parkinson's Detection & T10 & Dysarthria Detection & 0.30 \scriptsize{(0.17, 0.46)} & 0.48 \scriptsize{(0.37, 0.57)} & 0.58 \scriptsize{(0.38, 0.80)} & 0.49 \scriptsize{(0.35, 0.65)} & \textbf{0.60} \scriptsize{(0.48, 0.72)} & 0.55 \scriptsize{(0.43, 0.65)} & 0.56 \scriptsize{(0.49, 0.63)} & 0.55 \scriptsize{(0.44, 0.65)} & 0.58 \scriptsize{(0.50, 0.67)} & 0.40 \scriptsize{(0.30, 0.53)} & 0.47 \scriptsize{(0.39, 0.55)} & 0.56 \scriptsize{(0.47, 0.63)} \\
T13 & Parkinson's Detection & T17 & Disfluency Detection & 0.49 \scriptsize{(0.40, 0.58)} & 0.57 \scriptsize{(0.51, 0.64)} & 0.50 \scriptsize{(0.39, 0.61)} & 0.41 \scriptsize{(0.33, 0.50)} & 0.51 \scriptsize{(0.44, 0.56)} & 0.55 \scriptsize{(0.49, 0.60)} & 0.41 \scriptsize{(0.36, 0.45)} & 0.57 \scriptsize{(0.53, 0.62)} & 0.46 \scriptsize{(0.39, 0.55)} & \textbf{0.58} \scriptsize{(0.52, 0.64)} & 0.54 \scriptsize{(0.43, 0.63)} & 0.52 \scriptsize{(0.45, 0.58)} \\
T17 & Disfluency Detection & T13 & Parkinson's Detection & 0.29 \scriptsize{(0.15, 0.45)} & 0.50 \scriptsize{(0.32, 0.68)} & 0.57 \scriptsize{(0.42, 0.73)} & 0.34 \scriptsize{(0.19, 0.49)} & 0.37 \scriptsize{(0.17, 0.59)} & 0.47 \scriptsize{(0.32, 0.63)} & 0.37 \scriptsize{(0.27, 0.47)} & \textbf{0.76} \scriptsize{(0.60, 0.89)} & 0.44 \scriptsize{(0.24, 0.65)} & 0.53 \scriptsize{(0.43, 0.63)} & 0.58 \scriptsize{(0.42, 0.73)} & 0.55 \scriptsize{(0.41, 0.70)} \\
T17 & Disfluency Detection & T12 & Dysarthria Detection & 0.76 \scriptsize{(0.65, 0.85)} & 0.72 \scriptsize{(0.61, 0.83)} & 0.53 \scriptsize{(0.40, 0.64)} & 0.72 \scriptsize{(0.59, 0.83)} & 0.52 \scriptsize{(0.38, 0.67)} & 0.81 \scriptsize{(0.71, 0.89)} & 0.61 \scriptsize{(0.50, 0.71)} & 0.54 \scriptsize{(0.43, 0.65)} & 0.76 \scriptsize{(0.65, 0.85)} & 0.62 \scriptsize{(0.53, 0.70)} & 0.52 \scriptsize{(0.43, 0.62)} & \textbf{0.88} \scriptsize{(0.80, 0.95)} \\
T17 & Disfluency Detection & T10 & Dysarthria Detection & 0.66 \scriptsize{(0.55, 0.80)} & 0.73 \scriptsize{(0.59, 0.85)} & 0.57 \scriptsize{(0.46, 0.70)} & 0.66 \scriptsize{(0.50, 0.83)} & 0.37 \scriptsize{(0.19, 0.56)} & 0.69 \scriptsize{(0.60, 0.78)} & 0.74 \scriptsize{(0.61, 0.88)} & 0.69 \scriptsize{(0.53, 0.84)} & 0.48 \scriptsize{(0.33, 0.65)} & 0.62 \scriptsize{(0.54, 0.71)} & 0.64 \scriptsize{(0.55, 0.76)} & \textbf{0.74} \scriptsize{(0.61, 0.90)} \\
\midrule
\multicolumn{16}{l}{\textbf{Articulation (Phonatory / Respiratory)}} \\
\midrule
T24 & Respiratory Symptoms & T19 & Respiratory Symptoms & 0.54 \scriptsize{(0.53, 0.56)} & 0.55 \scriptsize{(0.54, 0.57)} & 0.53 \scriptsize{(0.52, 0.55)} & 0.50 \scriptsize{(0.49, 0.51)} & 0.52 \scriptsize{(0.51, 0.54)} & 0.58 \scriptsize{(0.57, 0.60)} & 0.55 \scriptsize{(0.54, 0.57)} & \textbf{0.61} \scriptsize{(0.60, 0.62)} & 0.49 \scriptsize{(0.48, 0.50)} & 0.53 \scriptsize{(0.51, 0.54)} & 0.54 \scriptsize{(0.53, 0.55)} & 0.51 \scriptsize{(0.50, 0.53)} \\
T27 & Vocal pathology detection & T19 & Respiratory Symptoms & 0.51 \scriptsize{(0.49, 0.52)} & 0.58 \scriptsize{(0.57, 0.59)} & 0.53 \scriptsize{(0.51, 0.54)} & 0.56 \scriptsize{(0.54, 0.57)} & 0.48 \scriptsize{(0.47, 0.50)} & \textbf{0.58} \scriptsize{(0.57, 0.60)} & 0.57 \scriptsize{(0.56, 0.59)} & 0.58 \scriptsize{(0.56, 0.59)} & 0.53 \scriptsize{(0.52, 0.55)} & 0.54 \scriptsize{(0.53, 0.56)} & 0.56 \scriptsize{(0.55, 0.58)} & 0.54 \scriptsize{(0.53, 0.56)} \\
T27 & Vocal pathology detection & T21 & COVID-19 Detection & 0.46 \scriptsize{(0.42, 0.50)} & 0.48 \scriptsize{(0.45, 0.52)} & 0.46 \scriptsize{(0.43, 0.50)} & 0.50 \scriptsize{(0.46, 0.53)} & 0.50 \scriptsize{(0.46, 0.55)} & \textbf{0.57} \scriptsize{(0.54, 0.61)} & 0.52 \scriptsize{(0.48, 0.56)} & 0.53 \scriptsize{(0.50, 0.57)} & 0.47 \scriptsize{(0.43, 0.51)} & 0.47 \scriptsize{(0.43, 0.51)} & 0.51 \scriptsize{(0.48, 0.55)} & 0.52 \scriptsize{(0.48, 0.56)} \\
T27 & Vocal pathology detection & T25 & COVID-19 Detection & 0.41 \scriptsize{(0.38, 0.43)} & 0.42 \scriptsize{(0.40, 0.45)} & 0.44 \scriptsize{(0.42, 0.46)} & 0.38 \scriptsize{(0.36, 0.41)} & 0.50 \scriptsize{(0.47, 0.52)} & 0.62 \scriptsize{(0.60, 0.64)} & \textbf{0.69} \scriptsize{(0.67, 0.71)} & 0.56 \scriptsize{(0.54, 0.58)} & 0.49 \scriptsize{(0.47, 0.51)} & 0.54 \scriptsize{(0.52, 0.56)} & 0.49 \scriptsize{(0.46, 0.51)} & 0.65 \scriptsize{(0.62, 0.67)} \\
T27 & Vocal pathology detection & T24 & Respiratory Symptoms & 0.42 \scriptsize{(0.40, 0.44)} & 0.46 \scriptsize{(0.44, 0.48)} & 0.48 \scriptsize{(0.46, 0.50)} & 0.42 \scriptsize{(0.40, 0.45)} & 0.47 \scriptsize{(0.45, 0.50)} & 0.61 \scriptsize{(0.59, 0.62)} & \textbf{0.69} \scriptsize{(0.67, 0.70)} & 0.59 \scriptsize{(0.57, 0.61)} & 0.50 \scriptsize{(0.49, 0.52)} & 0.58 \scriptsize{(0.56, 0.60)} & 0.51 \scriptsize{(0.49, 0.53)} & 0.65 \scriptsize{(0.63, 0.67)} \\
T19 & Respiratory Symptoms & T27 & Vocal pathology detection & 0.58 \scriptsize{(0.55, 0.61)} & 0.55 \scriptsize{(0.53, 0.58)} & 0.65 \scriptsize{(0.62, 0.68)} & 0.66 \scriptsize{(0.62, 0.69)} & 0.43 \scriptsize{(0.40, 0.47)} & 0.60 \scriptsize{(0.57, 0.62)} & 0.59 \scriptsize{(0.56, 0.61)} & 0.56 \scriptsize{(0.54, 0.59)} & 0.50 \scriptsize{(0.47, 0.54)} & 0.56 \scriptsize{(0.54, 0.59)} & \textbf{0.68} \scriptsize{(0.65, 0.71)} & 0.57 \scriptsize{(0.55, 0.59)} \\
T21 & COVID-19 Detection & T27 & Vocal pathology detection & 0.43 \scriptsize{(0.40, 0.45)} & 0.45 \scriptsize{(0.43, 0.47)} & 0.46 \scriptsize{(0.44, 0.49)} & 0.42 \scriptsize{(0.40, 0.45)} & 0.53 \scriptsize{(0.49, 0.57)} & \textbf{0.62} \scriptsize{(0.60, 0.65)} & 0.54 \scriptsize{(0.52, 0.56)} & 0.53 \scriptsize{(0.51, 0.55)} & 0.52 \scriptsize{(0.49, 0.56)} & 0.55 \scriptsize{(0.53, 0.58)} & 0.46 \scriptsize{(0.42, 0.49)} & 0.56 \scriptsize{(0.54, 0.58)} \\
T19 & Respiratory Symptoms & T24 & Respiratory Symptoms & 0.61 \scriptsize{(0.59, 0.63)} & \textbf{0.64} \scriptsize{(0.62, 0.66)} & 0.55 \scriptsize{(0.53, 0.57)} & 0.53 \scriptsize{(0.51, 0.55)} & 0.52 \scriptsize{(0.50, 0.54)} & 0.62 \scriptsize{(0.60, 0.64)} & 0.52 \scriptsize{(0.50, 0.53)} & 0.63 \scriptsize{(0.61, 0.65)} & 0.46 \scriptsize{(0.44, 0.48)} & 0.44 \scriptsize{(0.42, 0.46)} & 0.57 \scriptsize{(0.55, 0.59)} & 0.52 \scriptsize{(0.50, 0.54)} \\
T21 & COVID-19 Detection & T25 & COVID-19 Detection & 0.56 \scriptsize{(0.54, 0.59)} & \textbf{0.69} \scriptsize{(0.67, 0.71)} & 0.54 \scriptsize{(0.52, 0.56)} & 0.56 \scriptsize{(0.54, 0.58)} & 0.48 \scriptsize{(0.45, 0.50)} & 0.64 \scriptsize{(0.62, 0.66)} & 0.68 \scriptsize{(0.66, 0.70)} & 0.47 \scriptsize{(0.45, 0.49)} & 0.42 \scriptsize{(0.40, 0.44)} & 0.65 \scriptsize{(0.63, 0.67)} & 0.48 \scriptsize{(0.45, 0.50)} & 0.68 \scriptsize{(0.66, 0.70)} \\
T25 & COVID-19 Detection & T27 & Vocal pathology detection & 0.53 \scriptsize{(0.50, 0.56)} & 0.37 \scriptsize{(0.34, 0.39)} & 0.55 \scriptsize{(0.52, 0.58)} & 0.48 \scriptsize{(0.45, 0.51)} & 0.45 \scriptsize{(0.41, 0.48)} & \textbf{0.64} \scriptsize{(0.62, 0.67)} & 0.51 \scriptsize{(0.48, 0.52)} & 0.53 \scriptsize{(0.50, 0.55)} & 0.42 \scriptsize{(0.38, 0.45)} & 0.53 \scriptsize{(0.52, 0.55)} & 0.58 \scriptsize{(0.55, 0.61)} & 0.63 \scriptsize{(0.61, 0.66)} \\
T24 & Respiratory Symptoms & T27 & Vocal pathology detection & 0.47 \scriptsize{(0.44, 0.50)} & 0.44 \scriptsize{(0.42, 0.47)} & 0.58 \scriptsize{(0.55, 0.61)} & 0.52 \scriptsize{(0.50, 0.55)} & 0.42 \scriptsize{(0.38, 0.46)} & 0.63 \scriptsize{(0.60, 0.65)} & 0.57 \scriptsize{(0.54, 0.59)} & 0.57 \scriptsize{(0.55, 0.59)} & 0.49 \scriptsize{(0.46, 0.52)} & 0.58 \scriptsize{(0.57, 0.60)} & \textbf{0.64} \scriptsize{(0.62, 0.67)} & 0.63 \scriptsize{(0.61, 0.66)} \\
T25 & COVID-19 Detection & T21 & COVID-19 Detection & 0.54 \scriptsize{(0.50, 0.58)} & 0.61 \scriptsize{(0.57, 0.65)} & 0.55 \scriptsize{(0.51, 0.58)} & 0.51 \scriptsize{(0.47, 0.54)} & 0.48 \scriptsize{(0.44, 0.52)} & 0.57 \scriptsize{(0.53, 0.61)} & 0.57 \scriptsize{(0.52, 0.61)} & \textbf{0.61} \scriptsize{(0.57, 0.65)} & 0.51 \scriptsize{(0.47, 0.55)} & 0.57 \scriptsize{(0.53, 0.61)} & 0.55 \scriptsize{(0.51, 0.59)} & 0.54 \scriptsize{(0.50, 0.58)} \\
\midrule
\multicolumn{16}{l}{\textbf{Cross-category}} \\
\midrule
\multicolumn{2}{>{\raggedright\arraybackslash}p{0.18\linewidth}}{Conceptualization} & \multicolumn{2}{>{\raggedright\arraybackslash}p{0.18\linewidth}}{Formulation} & 0.66 \scriptsize{(0.60, 0.72)} & 0.49 \scriptsize{(0.43, 0.55)} & \textbf{0.67} \scriptsize{(0.61, 0.72)} & 0.33 \scriptsize{(0.29, 0.39)} & 0.30 \scriptsize{(0.26, 0.36)} & 0.61 \scriptsize{(0.57, 0.65)} & 0.27 \scriptsize{(0.22, 0.32)} & 0.60 \scriptsize{(0.54, 0.66)} & 0.55 \scriptsize{(0.48, 0.60)} & 0.26 \scriptsize{(0.21, 0.30)} & 0.51 \scriptsize{(0.45, 0.57)} & 0.27 \scriptsize{(0.23, 0.30)} \\
\multicolumn{2}{>{\raggedright\arraybackslash}p{0.18\linewidth}}{Conceptualization} & \multicolumn{2}{>{\raggedright\arraybackslash}p{0.18\linewidth}}{Articulation (Neuromuscular)} & 0.46 \scriptsize{(0.46, 0.47)} & 0.57 \scriptsize{(0.56, 0.57)} & 0.57 \scriptsize{(0.56, 0.57)} & \textbf{0.71} \scriptsize{(0.70, 0.71)} & 0.64 \scriptsize{(0.64, 0.65)} & 0.61 \scriptsize{(0.61, 0.62)} & 0.51 \scriptsize{(0.51, 0.52)} & 0.58 \scriptsize{(0.57, 0.59)} & 0.61 \scriptsize{(0.61, 0.62)} & 0.30 \scriptsize{(0.29, 0.31)} & 0.51 \scriptsize{(0.50, 0.51)} & 0.64 \scriptsize{(0.63, 0.64)} \\
\multicolumn{2}{>{\raggedright\arraybackslash}p{0.18\linewidth}}{Conceptualization} & \multicolumn{2}{>{\raggedright\arraybackslash}p{0.18\linewidth}}{Articulation (Phonatory / Respiratory)} & 0.48 \scriptsize{(0.46, 0.49)} & 0.50 \scriptsize{(0.49, 0.51)} & 0.55 \scriptsize{(0.54, 0.57)} & 0.48 \scriptsize{(0.46, 0.49)} & 0.48 \scriptsize{(0.47, 0.50)} & \textbf{0.60} \scriptsize{(0.59, 0.62)} & 0.49 \scriptsize{(0.48, 0.51)} & 0.45 \scriptsize{(0.44, 0.46)} & 0.50 \scriptsize{(0.49, 0.52)} & 0.44 \scriptsize{(0.43, 0.46)} & 0.45 \scriptsize{(0.44, 0.46)} & 0.55 \scriptsize{(0.54, 0.56)} \\
\multicolumn{2}{>{\raggedright\arraybackslash}p{0.18\linewidth}}{Formulation} & \multicolumn{2}{>{\raggedright\arraybackslash}p{0.18\linewidth}}{Conceptualization} & 0.51 \scriptsize{(0.49, 0.54)} & 0.46 \scriptsize{(0.44, 0.49)} & 0.52 \scriptsize{(0.49, 0.55)} & 0.53 \scriptsize{(0.51, 0.56)} & 0.44 \scriptsize{(0.42, 0.46)} & \textbf{0.74} \scriptsize{(0.71, 0.76)} & 0.48 \scriptsize{(0.47, 0.51)} & 0.53 \scriptsize{(0.51, 0.55)} & 0.52 \scriptsize{(0.50, 0.54)} & 0.49 \scriptsize{(0.47, 0.52)} & 0.47 \scriptsize{(0.45, 0.50)} & 0.45 \scriptsize{(0.43, 0.47)} \\
\multicolumn{2}{>{\raggedright\arraybackslash}p{0.18\linewidth}}{Formulation} & \multicolumn{2}{>{\raggedright\arraybackslash}p{0.18\linewidth}}{Articulation (Neuromuscular)} & 0.63 \scriptsize{(0.62, 0.63)} & 0.73 \scriptsize{(0.72, 0.73)} & \textbf{0.80} \scriptsize{(0.80, 0.81)} & 0.80 \scriptsize{(0.79, 0.80)} & 0.37 \scriptsize{(0.37, 0.38)} & 0.54 \scriptsize{(0.53, 0.54)} & 0.59 \scriptsize{(0.58, 0.60)} & 0.59 \scriptsize{(0.59, 0.60)} & 0.71 \scriptsize{(0.71, 0.72)} & 0.52 \scriptsize{(0.51, 0.52)} & 0.52 \scriptsize{(0.51, 0.53)} & 0.46 \scriptsize{(0.46, 0.47)} \\
\multicolumn{2}{>{\raggedright\arraybackslash}p{0.18\linewidth}}{Formulation} & \multicolumn{2}{>{\raggedright\arraybackslash}p{0.18\linewidth}}{Articulation (Phonatory / Respiratory)} & 0.51 \scriptsize{(0.50, 0.52)} & 0.53 \scriptsize{(0.52, 0.54)} & 0.58 \scriptsize{(0.57, 0.59)} & 0.53 \scriptsize{(0.52, 0.55)} & 0.53 \scriptsize{(0.51, 0.54)} & \textbf{0.58} \scriptsize{(0.57, 0.59)} & 0.54 \scriptsize{(0.53, 0.56)} & 0.53 \scriptsize{(0.52, 0.54)} & 0.54 \scriptsize{(0.53, 0.55)} & 0.52 \scriptsize{(0.51, 0.54)} & 0.54 \scriptsize{(0.53, 0.55)} & 0.49 \scriptsize{(0.48, 0.50)} \\
\multicolumn{2}{>{\raggedright\arraybackslash}p{0.18\linewidth}}{Articulation (Neuromuscular)} & \multicolumn{2}{>{\raggedright\arraybackslash}p{0.18\linewidth}}{Conceptualization} & 0.52 \scriptsize{(0.51, 0.53)} & 0.54 \scriptsize{(0.53, 0.55)} & 0.56 \scriptsize{(0.55, 0.58)} & \textbf{0.57} \scriptsize{(0.56, 0.58)} & 0.52 \scriptsize{(0.51, 0.54)} & 0.49 \scriptsize{(0.48, 0.50)} & 0.51 \scriptsize{(0.50, 0.52)} & 0.55 \scriptsize{(0.54, 0.56)} & 0.57 \scriptsize{(0.55, 0.58)} & 0.43 \scriptsize{(0.41, 0.44)} & 0.51 \scriptsize{(0.50, 0.52)} & 0.56 \scriptsize{(0.55, 0.57)} \\
\multicolumn{2}{>{\raggedright\arraybackslash}p{0.18\linewidth}}{Articulation (Neuromuscular)} & \multicolumn{2}{>{\raggedright\arraybackslash}p{0.18\linewidth}}{Formulation} & 0.81 \scriptsize{(0.79, 0.84)} & 0.86 \scriptsize{(0.84, 0.88)} & 0.73 \scriptsize{(0.70, 0.76)} & 0.48 \scriptsize{(0.45, 0.50)} & 0.36 \scriptsize{(0.33, 0.38)} & 0.51 \scriptsize{(0.48, 0.55)} & 0.86 \scriptsize{(0.85, 0.88)} & 0.82 \scriptsize{(0.80, 0.84)} & 0.73 \scriptsize{(0.70, 0.76)} & \textbf{0.87} \scriptsize{(0.86, 0.88)} & 0.82 \scriptsize{(0.80, 0.84)} & 0.76 \scriptsize{(0.74, 0.79)} \\
\multicolumn{2}{>{\raggedright\arraybackslash}p{0.18\linewidth}}{Articulation (Neuromuscular)} & \multicolumn{2}{>{\raggedright\arraybackslash}p{0.18\linewidth}}{Articulation (Phonatory / Respiratory)} & 0.53 \scriptsize{(0.52, 0.54)} & 0.54 \scriptsize{(0.53, 0.55)} & 0.55 \scriptsize{(0.54, 0.56)} & 0.53 \scriptsize{(0.53, 0.54)} & 0.51 \scriptsize{(0.50, 0.52)} & \textbf{0.59} \scriptsize{(0.58, 0.59)} & 0.55 \scriptsize{(0.54, 0.56)} & 0.53 \scriptsize{(0.52, 0.54)} & 0.53 \scriptsize{(0.52, 0.53)} & 0.55 \scriptsize{(0.55, 0.56)} & 0.56 \scriptsize{(0.55, 0.57)} & 0.54 \scriptsize{(0.54, 0.55)} \\
\multicolumn{2}{>{\raggedright\arraybackslash}p{0.18\linewidth}}{Articulation (Phonatory / Respiratory)} & \multicolumn{2}{>{\raggedright\arraybackslash}p{0.18\linewidth}}{Conceptualization} & 0.46 \scriptsize{(0.43, 0.48)} & 0.45 \scriptsize{(0.42, 0.49)} & 0.50 \scriptsize{(0.47, 0.55)} & 0.48 \scriptsize{(0.45, 0.51)} & 0.42 \scriptsize{(0.40, 0.44)} & \textbf{0.83} \scriptsize{(0.82, 0.85)} & 0.50 \scriptsize{(0.47, 0.54)} & 0.47 \scriptsize{(0.45, 0.51)} & 0.46 \scriptsize{(0.44, 0.48)} & 0.46 \scriptsize{(0.44, 0.50)} & 0.45 \scriptsize{(0.42, 0.48)} & 0.57 \scriptsize{(0.54, 0.63)} \\
\multicolumn{2}{>{\raggedright\arraybackslash}p{0.18\linewidth}}{Articulation (Phonatory / Respiratory)} & \multicolumn{2}{>{\raggedright\arraybackslash}p{0.18\linewidth}}{Formulation} & 0.64 \scriptsize{(0.59, 0.68)} & 0.79 \scriptsize{(0.75, 0.82)} & 0.73 \scriptsize{(0.68, 0.77)} & 0.80 \scriptsize{(0.75, 0.84)} & 0.67 \scriptsize{(0.61, 0.73)} & 0.61 \scriptsize{(0.57, 0.65)} & \textbf{0.88} \scriptsize{(0.85, 0.90)} & 0.73 \scriptsize{(0.69, 0.77)} & 0.55 \scriptsize{(0.49, 0.61)} & 0.81 \scriptsize{(0.77, 0.85)} & 0.57 \scriptsize{(0.51, 0.63)} & 0.86 \scriptsize{(0.83, 0.89)} \\
\multicolumn{2}{>{\raggedright\arraybackslash}p{0.18\linewidth}}{Articulation (Phonatory / Respiratory)} & \multicolumn{2}{>{\raggedright\arraybackslash}p{0.18\linewidth}}{Articulation (Neuromuscular)} & 0.63 \scriptsize{(0.63, 0.64)} & 0.59 \scriptsize{(0.59, 0.60)} & 0.67 \scriptsize{(0.67, 0.68)} & 0.72 \scriptsize{(0.71, 0.73)} & 0.52 \scriptsize{(0.51, 0.53)} & 0.68 \scriptsize{(0.68, 0.69)} & 0.75 \scriptsize{(0.74, 0.75)} & 0.55 \scriptsize{(0.54, 0.56)} & 0.53 \scriptsize{(0.53, 0.54)} & 0.65 \scriptsize{(0.64, 0.65)} & 0.59 \scriptsize{(0.58, 0.59)} & \textbf{0.75} \scriptsize{(0.74, 0.76)} \\
\bottomrule
\end{tabular}
}
\end{sidewaystable}


\newpage
\section*{NeurIPS Paper Checklist}

\begin{enumerate}

\item {\bf Claims}
    \item[] Question: Do the main claims made in the abstract and introduction accurately reflect the paper's contributions and scope?
    \item[] Answer: \answerYes{} 
    \item[] Justification: The abstract and introduction state two primary contributions: (1) \benchmarkname, a benchmark of 12 datasets and 27 tasks organized by the speech production mechanism, and (2) a systematic evaluation of 12 audio encoders in-domain and under zero-shot cross-condition transfer. Both contributions are realized in the paper. 

\item {\bf Limitations}
    \item[] Question: Does the paper discuss the limitations of the work performed by the authors?
    \item[] Answer: \answerYes{} 
    \item[] Justification: The paper includes a dedicated section (Section~\ref{sec:discussion}) discussing the limitations of this work. First, we acknowledge that the objectivity of clinical labels varies across the 27 tasks: many datasets compare healthy controls with patients who have lived with a condition for years rather than at the early-symptom stage relevant for screening, and several severity scores rely on self- or clinician-reported instruments subject to anchoring and recall bias. Second, the benchmark is primarily composed of English-language recordings and contains uneven representation of accents, ages, and gender, which limits generalization claims to broader populations and motivates demographic-stratified analyses as future work. Finally, the benchmark covers 13 datasets selected to span the four stages of speech production, but several clinically relevant conditions (e.g., Huntington's disease, pediatric speech sound disorders) are not represented because suitable public datasets are not yet available; the taxonomy is designed to accommodate such datasets as they become available. We frame these limitations as scoping decisions and concrete directions for future extension of the benchmark. Notably, despite these constraints, current state-of-the-art speech/audio models fail to reliably solve these foundational tasks, underscoring the need to establish robust performance on this benchmark before progressing to more complex real-world screening scenarios.
    

\item {\bf Theory assumptions and proofs}
    \item[] Question: For each theoretical result, does the paper provide the full set of assumptions and a complete (and correct) proof?
    \item[] Answer: \answerNA{} 
    \item[] Justification: We do not present theoretical results.

    \item {\bf Experimental result reproducibility}
    \item[] Question: Does the paper fully disclose all the information needed to reproduce the main experimental results of the paper to the extent that it affects the main claims and/or conclusions of the paper (regardless of whether the code and data are provided or not)?
    \item[] Answer: \answerYes{} 
    \item[] Justification: The paper provides comprehensive information to enable reproducibility of all experimental results. First, we release the complete benchmark codebase, including dataset processing pipelines, model training scripts, and evaluation protocols at \url{https://anonymous.4open.science/r/SpeechDx-F584}. Second, Section~\ref{sec:methods} and \autoref{app:implementation} provides detailed specifications of all model architectures, hyperparameters, training procedures, and data preprocessing steps. Third, for each of the 13 datasets in the benchmark, we document the train/validation/test splits, exclusion criteria, and task formulations in Section~\ref{sec:tasks-and-datasets}, \autoref{app:datasets} and \autoref{app:tasks}. For proprietary models (Whisper, HuBERT, etc.), we specify the exact model versions used in \autoref{app:models}. The combination of public data, released code, and detailed methodological documentation ensures that all main experimental results can be independently reproduced.
    

\item {\bf Open access to data and code}
    \item[] Question: Does the paper provide open access to the data and code, with sufficient instructions to faithfully reproduce the main experimental results, as described in supplemental material?
    \item[] Answer: \answerYes{}{} 
    \item[] Justification: We provide open access to both code and data with comprehensive reproduction instructions. The complete benchmark codebase is released at \url{https://anonymous.4open.science/r/SpeechDx-F584}, including: (1) data processing scripts for all 13 datasets, (2) preprocessing pipelines that transform raw audio recordings into train/validation/test splits matching those used in our experiments, (3) model training scripts with exact hyperparameters and environment specifications; (4) evaluation scripts that compute all reported metrics and generate tables/figures from the paper; and (5) a detailed README with instructions to reproduce each experimental result. All 13 datasets are publicly available from their original sources (see \autoref{tab:dataset-access}).
    

\item {\bf Experimental setting/details}
    \item[] Question: Does the paper specify all the training and test details (e.g., data splits, hyperparameters, how they were chosen, type of optimizer) necessary to understand the results?
    \item[] Answer: \answerYes{} 
    \item[] Justification: The paper provides comprehensive experimental details at multiple levels of granularity. Section~\ref{sec:methods} specifies the core experimental setup including training procedures, and evaluation protocols. For each dataset, we document the train/validation/test splits in Section~\ref{sec:tasks-and-datasets}, with sample counts provided in \autoref{tab:benchmark_tasks}. Hyperparameters selection and optimization is described in \autoref{app:implementation}. We further provide a benchmarking codebase that specifies additional information to reproduce our results at \url{https://anonymous.4open.science/r/SpeechDx-F584}.
    

\item {\bf Experiment statistical significance}
    \item[] Question: Does the paper report error bars suitably and correctly defined or other appropriate information about the statistical significance of the experiments?
    \item[] Answer: \answerYes{} 
    \item[] Justification: Given the scale of our benchmark (12 models evaluated across 27 tasks, with additional zero-shot evaluations), training each model multiple times with different random seeds is computationally prohibitive. Instead, we quantify uncertainty through bootstrap resampling of test set predictions, which captures variability due to finite sample size (refer to Section~\ref{sec:methods}. Confidence intervals are reported in all results tables and key figures. 
    

\item {\bf Experiments compute resources}
    \item[] Question: For each experiment, does the paper provide sufficient information on the computer resources (type of compute workers, memory, time of execution) needed to reproduce the experiments?
    \item[] Answer: \answerYes{} 
    \item[] Justification: We provide detailed compute resource information in Section~\ref{subsec:compute}.
    
\item {\bf Code of ethics}
    \item[] Question: Does the research conducted in the paper conform, in every respect, with the NeurIPS Code of Ethics \url{https://neurips.cc/public/EthicsGuidelines}?
    \item[] Answer: \answerYes{} 
    \item[] Justification: The research fully conforms to the NeurIPS Code of Ethics. We have carefully reviewed the guidelines and ensured compliance in all aspects. Our work uses only publicly available datasets with appropriate citations and respects original data usage licenses. All human subjects data comes from previously published datasets with proper IRB approval and informed consent obtained by the original data collectors.
    

\item {\bf Broader impacts}
    \item[] Question: Does the paper discuss both potential positive societal impacts and negative societal impacts of the work performed?
    \item[] Answer: \answerYes{} 
    \item[] Justification: The paper includes a discussion of broader impacts in Section~\ref{sec:discussion}. We discuss several positive impacts: improved accessibility of healthcare screening through low-cost voice-based tools, potential for early detection of respiratory and neurological conditions in underserved populations, and democratization of health monitoring through commodity devices. Regarding negative impacts, we note that while this is a research benchmark and not a system proposed for deployment, future models developed using this benchmark could face risks if deployed prematurely without clinical validation. We identify key concerns for such future work: demographic bias, privacy considerations, and the critical need for regulatory approval before any patient-facing applications. We explicitly caution that benchmark performance does not imply clinical readiness and recommend demographic-stratified evaluation as important future work.
    
    
\item {\bf Safeguards}
    \item[] Question: Does the paper describe safeguards that have been put in place for responsible release of data or models that have a high risk for misuse (e.g., pre-trained language models, image generators, or scraped datasets)?
    \item[] Answer: \answerNA{} 
    \item[] Justification: The benchmark does not release new datasets or pre-trained models with high misuse risk. We provide evaluation code and standardized task definitions for 13 existing public datasets, all of which have been previously released by their original authors with appropriate ethical review and data use agreements. We do not scrape new data from the Internet or release any trained models. The benchmark code includes documentation directing users to obtain dataset access through official channels where data use agreements are required.
    

\item {\bf Licenses for existing assets}
    \item[] Question: Are the creators or original owners of assets (e.g., code, data, models), used in the paper, properly credited and are the license and terms of use explicitly mentioned and properly respected?
    \item[] Answer: \answerYes{} 
    \item[] Justification: All datasets, models, and code libraries used in this work are properly credited with full citations to original papers. For datasets requiring data use agreements, we note the access procedure and direct users to the official sources (see \autoref{tab:dataset-access}).
    

\item {\bf New assets}
    \item[] Question: Are new assets introduced in the paper well documented and is the documentation provided alongside the assets?
    \item[] Answer: \answerYes{} 
    \item[] Justification: The benchmark codebase includes comprehensive documentation: a README with installation and usage instructions, detailed task specifications in Section~\ref{sec:tasks-and-datasets}, processing pipeline documentation for each dataset, and complete evaluation protocol descriptions in Section~\ref{sec:methods}. We do not collect new data; informed consent for all datasets was obtained by original data collectors as documented in their publications.
    

\item {\bf Crowdsourcing and research with human subjects}
    \item[] Question: For crowdsourcing experiments and research with human subjects, does the paper include the full text of instructions given to participants and screenshots, if applicable, as well as details about compensation (if any)? 
    \item[] Answer: \answerNA{} 
    \item[] Justification: The paper does not involve crowdsourcing or new human subjects research. All datasets are existing publicly available collections where human subjects protocols were handled by the original data collectors.

\item {\bf Institutional review board (IRB) approvals or equivalent for research with human subjects}
    \item[] Question: Does the paper describe potential risks incurred by study participants, whether such risks were disclosed to the subjects, and whether Institutional Review Board (IRB) approvals (or an equivalent approval/review based on the requirements of your country or institution) were obtained?
    \item[] Answer: \answerNA{} 
    \item[] Justification: The paper does not conduct new research with human subjects. All datasets are publicly available collections where IRB approval and informed consent were obtained by the original data collectors as documented in their respective publications.

\item {\bf Declaration of LLM usage}
    \item[] Question: Does the paper describe the usage of LLMs if it is an important, original, or non-standard component of the core methods in this research? Note that if the LLM is used only for writing, editing, or formatting purposes and does \emph{not} impact the core methodology, scientific rigor, or originality of the research, declaration is not required.
    \item[] Answer: \answerNA{} 
    \item[] Justification: With the exception of the audio encoders that we evaluated on our benchmark, which were written and released by their respective creators, our core experimental codebase was written entirely on our own by hand. LLMs were only used for peripheral affordances such as refactoring and documentation. 

\end{enumerate}

\end{document}